\pdfoutput=1

\documentclass[11pt]{article}

\usepackage[final]{acl}
\usepackage{booktabs}
\usepackage{tabularx}
\usepackage{multirow}
\usepackage{array}
\usepackage{times}
\usepackage{latexsym}
\usepackage{tabularx} 
\usepackage{booktabs} 
\usepackage{amssymb} 
\usepackage{float}

\usepackage[T1]{fontenc}

\usepackage[utf8]{inputenc}
\usepackage{booktabs}   
\usepackage{tabularx}   
\usepackage{makecell} 
\usepackage{array}      
\usepackage{subcaption}
\usepackage{microtype}
\usepackage{booktabs}
\usepackage[table]{xcolor} 

\usepackage{inconsolata}

\usepackage{graphicx}

\usepackage{algorithm}
\usepackage{algorithmic}
\usepackage{amsfonts} 
\usepackage{amssymb}
\usepackage{amsmath}
\usepackage{tabularx}
\usepackage{pifont}
\usepackage{pifont}
\usepackage{xcolor}
\definecolor{darkgreen}{rgb}{0.0, 0.5, 0.0} 
\usepackage[most]{tcolorbox} 

\usepackage{caption}
\title{Multimodal UNcommonsense: \\ From Odd to Ordinary and Ordinary to Odd}
\author{
\textbf{Yejin Son\textsuperscript{1}\thanks{Equal contribution}}, 
\textbf{Saejin Kim\textsuperscript{1}\footnotemark[1]}, 
\textbf{Dongjun Min\textsuperscript{1}}, 
\textbf{Youngjae Yu\textsuperscript{2}}
\\[3pt]
\textsuperscript{1}Yonsei University \quad
\textsuperscript{2}Seoul National University
\\[6pt]
\textbf{Correspondence:} \\
\texttt{yejinhand@yonsei.ac.kr}, \texttt{jerry0110@yonsei.ac.kr}, \texttt{youngjaeyu@snu.ac.kr}
}

\begin{document}
\maketitle
\begin{abstract}
Commonsense reasoning in multimodal contexts remains a foundational challenge in artificial intelligence. We introduce \textbf{Multimodal UNcommonsense (MUN)}, a 
 benchmark designed to evaluate models' ability to handle scenarios that deviate from typical visual or contextual expectations. MUN pairs visual scenes with surprising or unlikely outcomes described in natural language, prompting models to either rationalize seemingly odd images using everyday logic or uncover unexpected interpretations in ordinary scenes. To support this task, we propose a retrieval-based in-context learning (R-ICL) framework that transfers reasoning capabilities from larger models to smaller ones without additional training. Leveraging a novel \textbf{Multimodal Ensemble Retriever (MER)}, our method identifies semantically relevant exemplars even when image and text pairs are deliberately discordant. Experiments show an average improvement of 8.3\% over baseline ICL methods, highlighting the effectiveness of R-ICL in low-frequency, atypical settings. MUN opens new directions for evaluating and improving visual-language models’ robustness and adaptability in real-world, culturally diverse, and non-prototypical scenarios.

\end{abstract}


\begin{figure}[t]
\centering
\includegraphics[width=\columnwidth]{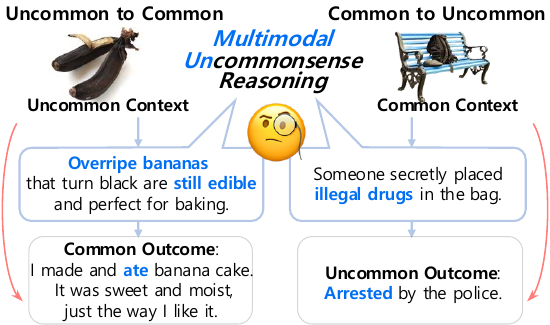}
\caption{Multimodal UNcommonsense Reasoning aims to produce explanations that make given outcomes appear likely.  For example, overripe bananas (an uncommon context) can still be used for baking a sweet, moist banana cake (a common outcome), while a bag on a bench (common context) leads to an arrest (uncommon outcome). This highlights the challenge of bridging visual cues with logical reasoning, as addressed in our Multimodal Uncommonsense (MUN) dataset.}
\label{fig:teasure_figure}
\end{figure}

\begin{figure*}[t]
\centering
\includegraphics[width=1.5\columnwidth]{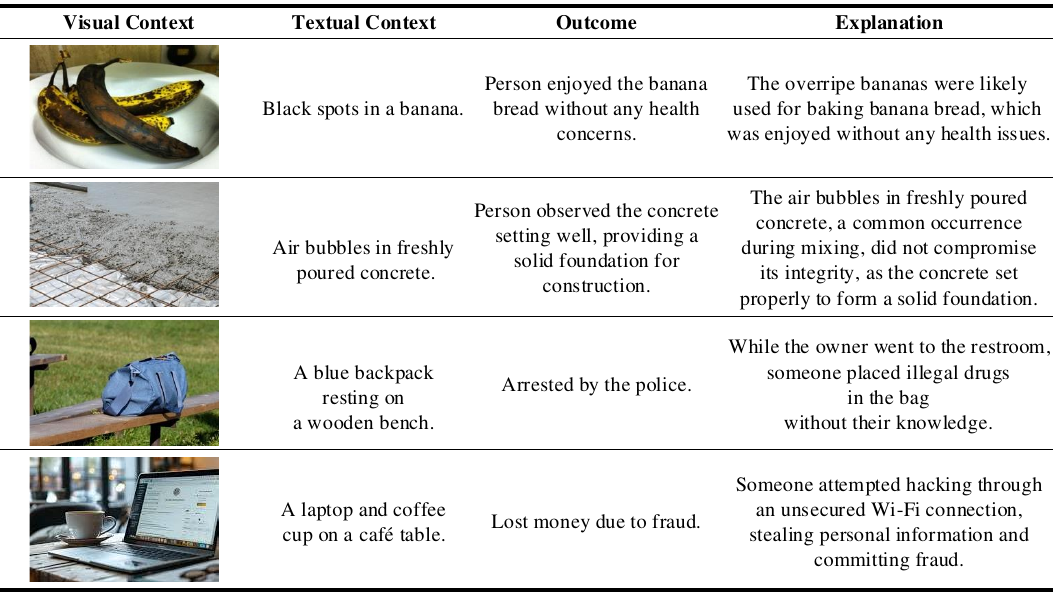}
 \caption{MUN examples. The first two examples are from MUN-vis and the next two examples
come from MUN-lang; explanations are written by human annotators. Note that textual context is only used during dataset generation.}
\label{fig:MUN_vis_ex}
\end{figure*}

\section{Introduction}
In everyday life, commonsense functions as an invisible framework, akin to "dark matter" in the universe. Though we cannot directly perceive it, commonsense subtly influences our decisions, such as recognizing social norms or interpreting ambiguous situations~\cite{bosselut2019comet, tao2024cultural}. 

While this commonsense often leads to stable reasoning in familiar contexts, it can falter when confronted with visual or textual cues that fall outside typical experience or cultural familiarity~\cite{wang2023commonsensevis}. 
However, most existing benchmarks for commonsense reasoning evaluate models on frequent or prototypical cases that are well-covered by large-scale English-language corpora and standard vision-language datasets~\cite{brown2020language, raffel2020exploring, hendricks2016generating, agrawal2017cvt, li2019visualbert}.
As a result, current AI systems exhibit significant brittleness when faced with rare, ambiguous, or culturally-specific phenomena that lie beyond the training distribution.

To address this gap, we introduce the \textbf{Multimodal UNcommonsense (MUN)} benchmark: a human-curated dataset specifically designed to evaluate models’ ability to reason about \textbf{uncommon or counterintuitive outcomes} in visual contexts. Unlike prior datasets that emphasize prototypical commonsense, MUN centers on visual situations that violate typical expectations, such as an overripe banana being preferable for baking, or culturally specific gestures like the Indian head wobble indicating agreement. These cases challenge the model to reconcile visual oddities with logically or culturally grounded explanations.

The necessity of MUN lies in its focus on \textbf{low-frequency, multimodal reasoning}, a facet critical for real-world applications where visual inputs and commonsense expectations often diverge. By constructing and evaluating against such examples, MUN serves as a benchmark that complements existing datasets, expanding the scope of commonsense evaluation beyond conventional boundaries.

We collected human-written and LLM-generated explanations for each case, revealing a significant gap in interpretability and diversity. While LLM explanations are often precise, human annotations offer a broader range of perspectives, as noted in prior work~\cite{zhao2023uncommonsense}. We leverage both via augmentation to build a high-quality benchmark that supports rich supervision and evaluation.

To enhance reasoning in visually and contextually atypical scenarios, we adopt a retrieval-based in-context learning (R-ICL) method~\cite{lin2022unsupervised,lin2023unlocking}, which improves smaller models by leveraging semantically relevant exemplars generated by larger models.
Specifically, to effectively identify these exemplars within scenarios exhibiting visual-textual discordance, we introduce an innovative retrieval framework known as the \textit{Multimodal Ensemble Retriever} (MER).
MER independently scores similarity in each modality and fuses them via a tunable weighting mechanism, enabling flexible retrieval in the presence of intentionally discordant image-text pairs in MUN.

Unlike conventional retrievers that assume strong cross-modal alignment, MER accommodates the unaligned nature of our benchmark. To the best of our knowledge, this is the first application of R-ICL in a setting where visual and textual signals are deliberately discordant, enabling abductive reasoning over unaligned multimodal inputs. This approach yields an average 8.3\% increase in win rate over a random baseline, demonstrating the effectiveness of R-ICL in boosting nuanced multimodal reasoning.

By connecting intuitive impressions with underlying truths in visually uncommon scenarios, MUN lays the groundwork for building trustworthy AI systems capable of reasoning beyond the obvious across cultures, contexts, and expectations.

\section{Related Work}

\paragraph{Abductive Reasoning.} Abductive reasoning, central to commonsense, involves inferring the most plausible explanations from incomplete observations. While various efforts have explored textual and multimodal abductive reasoning, each existing approach exhibits limitations (Table \ref{tab:comparison}). For example, Abductive-NLI \cite{bhagavatula2019abductive} focuses solely on textual input in everyday scenarios without visual grounding. Sherlock~\cite{hessel2022abduction} integrates real images and text but remains constrained to common situations and unidirectional reasoning. UNcommonsense \cite{zhao2023uncommonsense} targets uncommon contexts but lacks visual signals, while NL-Eye \cite{ventura2024nl} employs synthetic images without adequately addressing non-commonsensical scenarios or bidirectionality. As a result, none of these existing approaches simultaneously incorporate real imagery, handle uncommon contexts, and support bidirectional abductive inference. In contrast, our proposed MUN (Multimodal UNcommonsense) dataset integrates real images and text, actively considers uncommon scenarios, and enables bidirectional reasoning, thereby addressing these gaps and offering a more comprehensive and nuanced abductive reasoning benchmark.

\begin{table}
\centering
\scriptsize 
\renewcommand{\arraystretch}{1.2} 
\begin{tabularx}{\columnwidth}{>{\centering\arraybackslash}m{1.7cm}>{\centering\arraybackslash}m{0.8cm}>{\centering\arraybackslash}m{0.7cm}>{\centering\arraybackslash}m{0.9cm}>{\centering\arraybackslash}m{1.6cm}}
\toprule
\textbf{Dataset} & \textbf{Modality} & \textbf{Real image} & \textbf{Uncommon} & \textbf{Bi-direction Reasoning?}\\
\midrule
Abductive-NLI & T & {\color{red}\ding{55}} & {~~~~~\color{red}\ding{55}} & {\color{red}\ding{55}} \\
Sherlock & I+T & {\color{darkgreen}\ding{51}} & {~~~~~\color{red}\ding{55}} & {\color{red}\ding{55}} \\
Uncommonsense & T & {\color{red}\ding{55}} & {~~~~~\color{darkgreen}\ding{51}} & {\color{red}\ding{55}} \\
NL-Eye & I+T & {\color{red}\ding{55}} & {~~~~~\color{red}\ding{55}} & {\color{red}\ding{55}} \\
\midrule
\textbf{MUN (Ours)} & I+T & {\color{darkgreen}\ding{51}} & {~~~~~\color{darkgreen}\ding{51}} & {\color{darkgreen}\ding{51}} \\
\bottomrule
\end{tabularx}
\vspace{-1mm} 
\caption{Comparison with Abductive reasoning benchmark. \cite{bhagavatula2019abductive, hessel2022abduction,zhao2023uncommonsense,ventura2024nl} "I" stands for Image and "T" stands for Text. The MUN uniquely supports "Bi-direction UNcommonsense Reasoning," combining unusual contexts, outcomes, and nuanced visual scenarios.}
\label{tab:comparison}
\end{table}

\paragraph{Retrieval-Augmented and In-Context Learning}
Recent advances in LLMs~\cite{brown2020language, chowdhery2022palm, achiam2023gpt} and VLMs~\cite{alayrac2022flamingo, li2023blip2} have shown remarkable zero- and few-shot learning capabilities. However, their reasoning often remains tied to patterns entrenched in their training data. Retrieval-augmented paradigms~\cite{thoppilan2022lamda} and in-context learning (ICL) techniques~\cite{wei2022chain, zhou2022least} represent promising strategies to extend model capabilities beyond memorized knowledge. By dynamically incorporating external documents, exemplars, or contextual cues, models can handle more complex reasoning tasks and adapt to new domains. In visual domains, multi-source retrieval \cite{zhu2020dark, shao2024cpt} and retrieval-based image grounding show potential. Our work aligns with this trend by using a retrieval-based ICL approach. We retrieve both textual and visual exemplars from MUN scenarios, guiding model reasoning and distilling complex abductive and cultural logic into accessible formats. This approach assists smaller VLMs in navigating unusual scenarios and producing coherent, contextually rich explanations.

\section{Multimodal UNcommonsense (MUN) }
 To advance research in Visual Uncommonsense Reasoning, we have constructed the benchmark \textbf{Multimodal UNcommonsense (MUN)}, created to challenge models with scenarios that diverge from standard visual or contextual expectations. Inspired by prior work~\cite{zhao2023uncommonsense} on uncommonsense reasoning, specifically abductive reasoning about unusual situations, our dataset adopts a structured \textit{context-results-explanation} paradigm. In this framework, models are required to interpret an image-based context along with a textual scenario (the results) and then generate an explanation that reconciles the two.

\subsection{Task Settings}

We focus on two complementary task settings that emphasize the delicate interplay between visual cues and textual reasoning.

\paragraph{MUN-vis: Uncommon Image (Context) → Common Results}
In this task, the model is presented with an image that initially appears visually peculiar or “uncommon,” representing situations that occur with low frequency or probability. Despite this apparent strangeness, the goal is to generate a coherent explanation that normalizes the scenario and demonstrates that it is actually common or perfectly reasonable. For instance, in the first row of Figure \ref{fig:MUN_vis_ex}, a photograph of a blackened banana might initially seem unusual. However, the outcome states, "Person enjoyed the banana bread without any health concerns," indicating that an explanation such as "The bananas were overripe and therefore used for baking banana bread, which was enjoyed without any health issues" is needed to bridge the gap between the context and the outcome. This task involves generating explanations that connect seemingly peculiar visual inputs to familiar and logical everyday contexts.
\paragraph{MUN-lang: Common Image (Context) → Uncommon Results}
In this scenario, the model is presented with an image that appears completely ordinary but must explain an unusual or "uncommon" textual outcome associated with it.  
In Figure \ref{fig:MUN_vis_ex}, the context depicted in the third row shows a seemingly typical scene of a blue backpack resting on a wooden bench, while the outcome is "Arrested by the police," which does not naturally align with the given context. 
The explanation must bridge this gap by uncovering less obvious details, such as "While the owner was in the restroom, someone secretly placed illegal drugs in the bag without their knowledge," providing a surprising yet plausible rationale to make sense of the discordant situation.

\subsection{Dataset Creation}

We constructed the MUN dataset through a multi-step process, generating diverse “uncommonsense” scenarios that challenge multimodal reasoning models.

\paragraph{Scenario Generation}
We used GPT-4o to produce a diverse range of textual scenarios(contexts). For MUN-vis, we instructed the model to depict scenes initially appearing visually odd but ultimately normal. For MUN-lang, we asked for ordinary-looking scenes that conceal surprising rationales.
Our prompting strategy encouraged the model to analyze hypothetical image-text pairs, classify them as “normal” or “anomalous,” and provide brief explanations. By varying visual and contextual cues and highlighting underlying reasons, we obtained scenarios rich in cultural context, sensory detail, and conceptual twists.
This approach guided GPT-4o to produce structured, logically grounded explanations. In MUN-vis entries, seemingly strange images were normalized by uncovering rational backstories. In MUN-lang entries, mundane appearances were reinterpreted through hidden surprises or unconventional practices.

\paragraph{Filtering for Ensuring Diversity}
To ensure a diverse dataset, we implemented a comprehensive filtering process after generating a large pool of candidate scenarios. Observing numerous similar scenarios, we prioritized removing them to promote diversity and minimize redundancy. Using the Dedupe library,\footnote{\url{https://github.com/dedupeio/dedupe}.} A specialized tool for data deduplication, we effectively eliminated duplicates.

Inspired by diversity filtering~\cite{han-etal-2023-reading}, we further enriched the diversity of contexts by identifying a list of specific keywords. Examples were filtered out if the language description of an image contained any of these keywords. To maintain balance, we ensured that the occurrence of these keywords in the contexts remained below 20.

\paragraph{Image Pairing and Selection Process}
For each textual scenario(context), we first retrieved five candidate images using the Bing Web Search API\footnote{The Bing Web Search API:\url{https://www.microsoft.com/en-us/bing/apis/bing-web-search-api}.}, then manually reviewed them to select the image that best reflected the scenario’s uniqueness or ordinariness. If suitable images were not found through automated searches, we conducted additional manual searches to identify appropriate options.

By incorporating real-world images, the model can achieve more stable and generalizable reasoning capabilities, as demonstrated by research on ALBEF~\cite{li2021align}, BLIP~\cite{li2022blip}, and LLaVA~\cite{ liu2023llava}, as well as large-scale, diverse image resources like LAION-5B~\cite{schuhmann2022laion}. Building datasets grounded in authentic visuals enables expansion to cover rare situations and cultural nuances. Through iterative refinement, this approach surpasses existing limitations and supports more nuanced cross-domain reasoning.

\paragraph{Human Explanation Generation}  
We recruited 26 graduate students specializing in computer science and artificial intelligence as annotators to participate in the primary explanation-writing tasks. All participants were proficient in English, and the interface and instructions were provided in English. To ensure a fair and efficient workflow, the tasks were divided into small batches, with the workload evenly distributed among the annotators. This approach prevented any single annotator from being overburdened, thereby maintaining the consistency and quality of the dataset.
Additionally, to enhance the contextual reliability of the dataset, annotators were instructed not to write explanations for scenarios they deemed irrelevant or inappropriate. This measure prevented the inclusion of unnecessary or non-essential explanations. Furthermore, annotators were encouraged to logically infer and articulate the reasons behind outcomes that appeared mismatched within the provided visual context.

\paragraph{LLM-Enhanced Human-Written Explanations}  
As shown in subsequent analysis, and consistent with previous studies~\cite{zhao2023uncommonsense}, human-written explanations demonstrate the diversity and broad understanding, while LLM-generated responses tend to be relatively narrow and specific. We aim to combine these complementary strengths to further refine human annotations. Specifically, we use carefully crafted prompts to guide GPT-4o in improving human-written explanations, enabling it to present clearer and more specific logical connections between visual scenarios and the given uncommon outcomes. This process preserves the diversity and nuance of human explanations while leveraging the precision of LLMs, resulting in an improved set of explanations that provide a more informative baseline for comparison.

\subsection{Data Analysis}
The MUN dataset includes two subtasks: MUN-vis with 515 instances of visually uncommon contexts and common outcomes, and MUN-lang with 500 instances of visually common contexts and uncommon outcomes, totaling 1,015 visual context-outcome pairs. Human explanations were collected for 143 instances from MUN-vis and 156 from MUN-lang, with LLM-generated explanations for all pairs.

\begin{figure*}[t]
\centering
\includegraphics[scale=0.35]{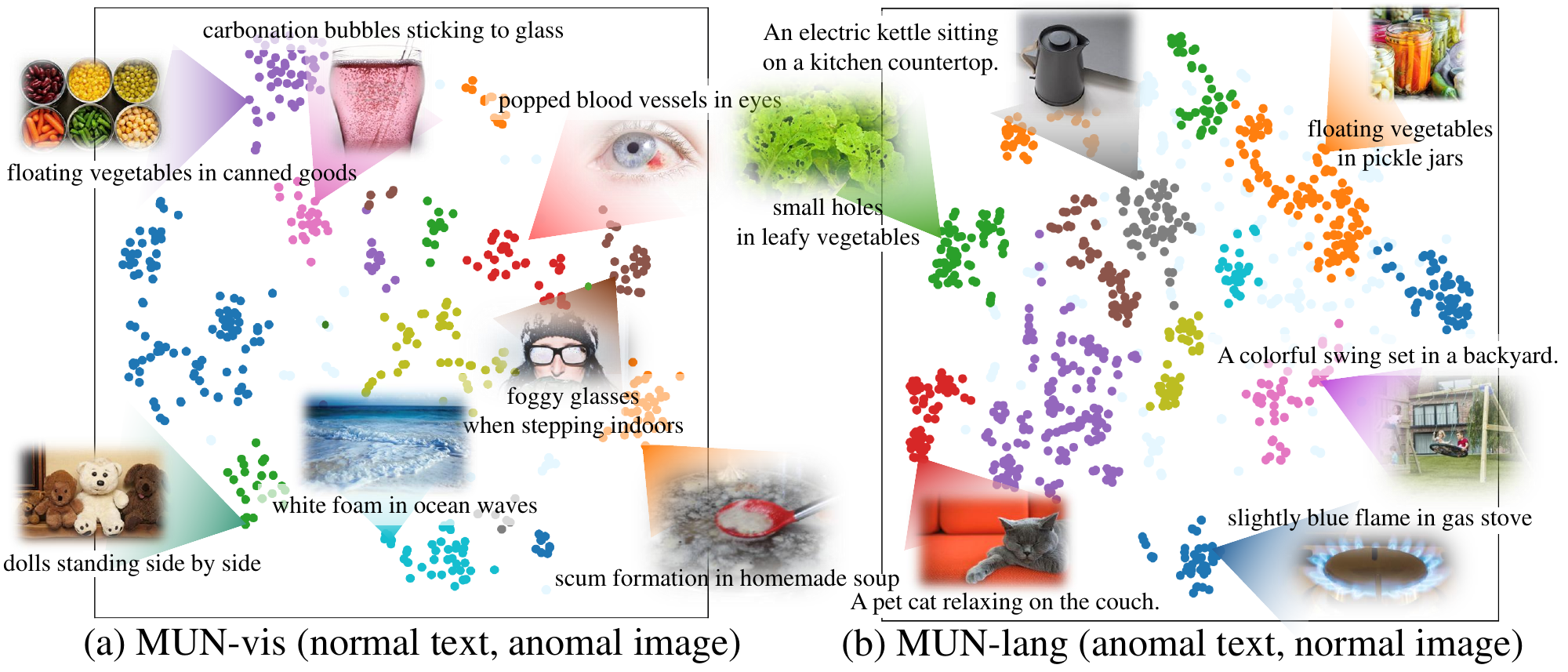}
\vspace{-2mm} 
\caption{
t-SNE visualization of MUN-vis (a) and MUN-lang (b) based SimCSE~\cite{gao-etal-2021-simcse} across categories.
}
\label{fig:TSNE}
\end{figure*}

\paragraph{Diversity of MUN}
The MUN dataset spans a broad range of scenarios across various categories, with each example including detailed textual explanations linking visual context to outcomes. While certain categories may be emphasized, individual examples still capture complex, multilayered scenes. For detailed reports on the frequencies of topics and their combinations, see Appendix~\ref{sec:dataset_categories}. 
The t-SNE~\cite{maaten2008visualizing} visualization (Figure \ref{fig:TSNE}) reveals that textual contexts cluster into distinct groups, covering a wide array of subjects.

\begin{figure}[t]
\centering
\includegraphics[scale=0.67]{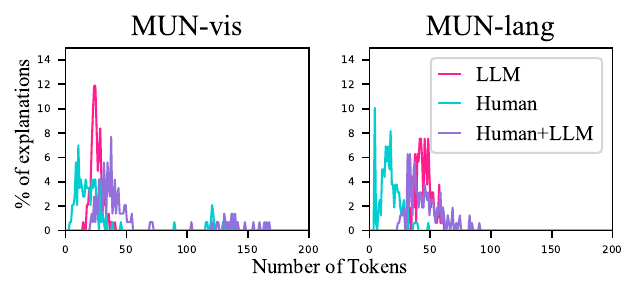}
\vspace{-2mm} 
\caption{Explanation token length distributions in MUN: The left section represents MUN-vis, while the right section depicts MUN-lang, derived from calculations on the development sets of each data subset.}
\label{fig:explanation_length_dist}
\end{figure}

\begin{figure}[t]
\centering
\includegraphics[scale=0.6]{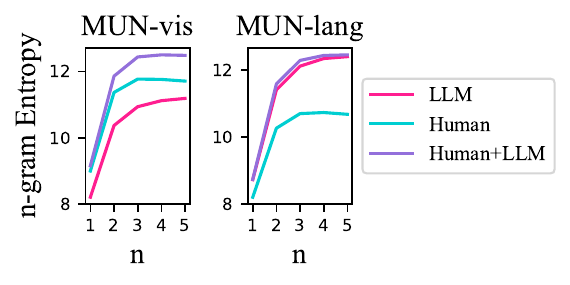}
\caption{The n-gram distribution entropies for MUN-vis (left) and MUN-lang (right) were calculated based on the development sets for each data subset.}
\vspace{-4mm} 
\label{fig:ngram_diversity}
\end{figure}

\begin{table}[t]
\centering
\scriptsize 
\renewcommand{\arraystretch}{0.8} 
\resizebox{0.7\columnwidth}{!}{ 
\begin{tabular}{lccc}
\toprule
~LV & Human(\%) & LLM(\%) & Human+\tiny{LLM}(\%) \\
\midrule
~~1 & 30.5 & 0.3 & 1.3 \\
~~2 & 40.1 & 8.9 & 9.3 \\
~~3 & 8.6 & 35.4 & 20.5 \\
~~4 & 11.9 & 55.0 & 62.6 \\
~~5 & 8.9 & 0.3 & 6.3 \\
\midrule
Avg. & \textbf{2.29} & \textbf{3.46} & \textbf{3.63} \\
\bottomrule
\end{tabular}
}
\vspace{-1mm} 
\caption{
Comparison of the specificity of explanations written by humans (Human), explanations generated by LLMs (LLM), and human-written explanations enhanced by LLMs (LLM+Human). Each value in the table represents the proportion of explanations rated at each specificity level (1 to 5) in percentile.}
\label{tab:specificty_comparison}
\end{table}






\subsection{Comparison Analysis of Explanations}

Consistent with Uncommonsense~\cite{zhao2023uncommonsense}, there were noticeable differences in both the length and lexical diversity of explanations generated by LLM, Human, and Human+LLM. 
Figure~\ref{fig:explanation_length_dist} illustrates the distribution of explanation lengths.
In the \textbf{MUN-vis} task, human explanations were relatively long and variable, averaging $32.0 \pm 38.5$ tokens. LLM explanations, on the other hand, were shorter and more stable at $25.1 \pm 4.4$ tokens, whereas Human+LLM explanations were longer at $50.7 \pm 37.2$ tokens, offering more detailed content. In the \textbf{MUN-lang} task, humans produced shorter explanations ($16.3 \pm 7.9$ tokens), while LLM outputs were longer and more consistent ($44.5 \pm 6.4$ tokens). This pattern suggests that, in more open-ended tasks like MUN-lang, LLMs produce richer and longer explanations, whereas in more structured tasks like MUN-vis, humans tend to provide longer descriptions. Human+LLM explanations reached $44.6 \pm 13.1$ tokens, approaching LLM-level length while combining human creativity with LLM stability.

To quantify lexical diversity, we measured $n$-gram entropy ($n \in \{1,\dots,5\}$) as shown in Figure~\ref{fig:ngram_diversity}, conducting 1,000 bootstrap iterations\footnote{In each iteration, one explanation was randomly selected per context-outcome pair from each subset.}. In \textbf{MUN-vis}, human explanations displayed higher $n$-gram entropy than LLM explanations, and Human+LLM exceeded human entropy, reflecting a synergy where human variability and LLM precision were combined. LLM explanations showed lower entropy, possibly due to the task’s structured nature. In \textbf{MUN-lang}, LLM entropy was similar to or even higher than Human+LLM’s and significantly exceeded that of humans, indicating that LLMs employ more diverse wording in open-ended tasks, whereas human language use is more constrained. Human+LLM still maintained high entropy, effectively blending human creativity and LLM rigor.

Recent work~\citep{Stiennon2020,liu-etal-2023-g} suggests that LLMs can reliably evaluate qualitative aspects of text, such as specificity, given well-structured prompts. Following this approach, we employed GPT-4o to evaluate the specificity (scores 1 to 5).

Table \ref{tab:specificty_comparison} shows that human explanations had a high proportion (70.6\%) of low specificity (scores 1 to 2) and a relatively low proportion (20.8\%) of high specificity (scores 4 to 5). LLM explanations generally maintained moderate to high specificity (scores 3 to 4), with a large proportion of 4-point ratings (55.0\%), but very few achieved the highest specificity (0.3\% for score 5). In contrast, Human+LLM had an even higher proportion of 4-point ratings (62.6\%) and improved the proportion of 5-point ratings (6.3\%), thereby maximizing overall specificity. This demonstrates that LLMs can refine and expand upon human input, achieving a higher level of detail and specificity, and that a Human+LLM approach can combine the strengths of both while compensating for their respective weaknesses. Based on these results, Human+LLM was deemed to be the best and was selected as the baseline for evaluation.\footnote{Further analysis of how humans perceive differences between human-written and LLM-generated or LLM-augmented responses is provided in the Appendix~\ref{sec:human_agreement_comparison}.} 

\section{Visual Uncommonsense Reasoning with Retrieved In-Context Learning}

The core aim of the MUN dataset is to challenge models with atypical, low-frequency visual-text scenarios that resist conventional commonsense interpretations. Unlike standard benchmarks, MUN focuses on “uncommonsense” reasoning, where the model must infer nuanced, often abductive rationales for unusual outcomes. This pushes beyond the straightforward pattern matching that large vision-language models (VLMs) typically excel at due to their extensive pretraining on statistically dominant patterns.

However, when confronted with such atypical scenes, models tend to regress to high-probability patterns learned during pretraining, producing literal or overly generic captions rather than abductive explanations~\cite{hessel2022abduction}. To enable more contextually grounded reasoning in these unfamiliar scenarios, we adopt retrieval-based in-context learning (ICL) to surface relevant yet semantically non-obvious examples.

Conventional retrieval methods, which assume strong alignment between modalities, often struggle with the intentional divergence of image-text pairs in MUN. To address this, we introduce a \textit{Multimodal Ensemble Retriever} (MER) that scores image and text similarities separately and combines them via a tunable fusion mechanism. This approach enables MER to retrieve semantically coherent examples even when the visual and textual cues signal distinct or conflicting commonsense expectations.

Specifically, MER embeds (image, text) pairs using a CLIP-style image encoder and a BERT-based text encoder, and computes cosine similarity between the query and dataset entries for each modality. The two similarity scores are then integrated using a weighting coefficient $\alpha$ that balances the contribution of each modality. This separate-but-aligned retrieval strategy allows MER to flexibly accommodate modality-specific signals, providing a principled mechanism for bridging conceptual gaps in visually grounded abductive reasoning.

To the best of our knowledge, this is the first application of such a dual-scoring retrieval framework in a setting where the visual and textual modalities are intentionally discordant. The full formulation and algorithmic implementation are provided in Appendix~\ref{sec:retrieval_algo}.

\section{Experiments}
\label{sec:experiments}

\begin{table}[!t]
\centering
\resizebox{\linewidth}{!}{
\begin{tabular}{p{0.85cm} l | c | c c | c c | c c }
\toprule
\multirow{2}{*}{\textbf{Dataset}} & \multirow{2}{*}{\textbf{~~~Model}} &
\multirow{2}{*}{\textbf{0-shot}} &
\multicolumn{2}{c|}{\textbf{1-shot}} &
\multicolumn{2}{c|}{\textbf{3-shot}} &
\multicolumn{2}{c}{\textbf{5-shot}} \\
\cmidrule(lr){4-5}\cmidrule(lr){6-7}\cmidrule(lr){8-9}
& &  & Rand. & R-ICL & 
           Rand. & R-ICL & 
           Rand. & R-ICL \\
\midrule
\multirow{8}{*}{\makecell[c]{\textbf{MUN}\\\textbf{vis}}}
& \textit{Gemma}3       & 0.335 & 0.428 & 0.457& 0.422 & \textbf{0.491}& 0.393 & 0.382\\
& \textit{InternVL }2.5 & 0.243 & 0.121 & 0.254 & 0.208 & 0.312& 0.387 & \textbf{0.434} \\
& \textit{LLaVA OV}     & 0.301 & 0.439 & 0.405 & 0.434 & 0.376 & \textbf{0.474} & 0.445\\
& \textit{Phi }3.5\textit{v}     & 0.283 & 0.324 & 0.335 & 0.387 & 0.387 & 0.428 & \textbf{0.445}\\
& \textit{Phi }4\textit{mm}      & 0.410 & 0.312 & 0.272 & 0.393 & 0.387 & \textbf{0.630} & 0.618 \\
& \textit{Qwen2.5}\textit{ VL}   & 0.364 & 0.428 & 0.387 & 0.422 & 0.428 & 0.439 & \textbf{0.538}\\
& \textit{Qwen}2\textit{ VL}     & 0.225 & 0.399 & 0.405& 0.283 & 0.376& 0.272 & \textbf{0.486}\\
\midrule
\multirow{8}{*}{\makecell[c]{\textbf{MUN}\\\textbf{lang}}}
& \textit{Gemma}3       & 0.257 & 0.341 & 0.418 & 0.430 & 0.454 & 0.498 & \textbf{0.546}\\
& \textit{InternVL }2.5 & 0.325 & 0.273 & 0.293 & 0.361 & 0.357 & 0.430 & \textbf{ 0.470}\\
& \textit{LLaVA OV}     & 0.285 & 0.301 & 0.325 & 0.333 & 0.369 & 0.365 & \textbf{0.369} \\
& \textit{Phi }3.5\textit{v}     & 0.337 & 0.353 & 0.329 & 0.410 & 0.390 & 0.430 & \textbf{0.442} \\
& \textit{Phi }4\textit{mm}      & 0.357 & 0.502 & 0.582 & 0.534 & 0.554  & 0.651 & \textbf{0.655}  \\
& \textit{Qwen2.5}\textit{ VL}   & 0.422 & 0.353 & 0.329  & 0.357 & 0.357  & 0.410 & \textbf{0.426} \\
& \textit{Qwen}2\textit{ VL}     & 0.349 & 0.349 & 0.321 & 0.365 & 0.341 &  \textbf{0.365} & 0.357 \\
\bottomrule
\end{tabular}}
\vspace{-2mm}
\caption{Comparison of models in different shot settings, measured by winning ratio against human-assisted explanations(higher is better). "Random" indicates randomly chosen examples, and "R-ICL" indicates retrieved examples for in-context learning. Model outputs were compared with Human+LLM explanations, judged using LLM. }
\label{tab:combined_results}
\end{table}

We evaluate the effectiveness of our proposed retrieved in-context learning (ICL) approach for multimodal uncommonsense reasoning using the MUN dataset.
Our experimental study is organised to shed light on two core questions:
\begin{list}{}{
  \setlength{\leftmargin}{1cm}
  \setlength{\labelwidth}{1cm}
  \setlength{\topsep}{1pt}
  \setlength{\partopsep}{1pt}
  \setlength{\itemsep}{1pt}
  \setlength{\parsep}{1pt}
}
  \item[\textbf{RQ1.}] How does the number of in-context examples (shots) affect model performance?
  \item[\textbf{RQ2.}] What is the impact of retrieval-based examples compared to randomly selected ones?
\end{list}
To establish robust baselines and ensure comprehensive evaluation, we benchmark several state-of-the-art vision-language models (VLMs) and utilize a multimodal ensemble retriever for our retrieval-based ICL approach. 

\subsection{Models Selection and Retrieval Mechanism}
\label{ssec:models}

We evaluate seven interleaved VLMs spanning different architecture families and size scales:
Qwen2-VL~\cite{wang2024qwen2}, Qwen2.5-VL~\cite{bai2025qwen2_5vl},  Phi-3.5-vision~\cite{abdin2024phi}, Phi-4-multimodal~\cite{abouelenin2025phi4}, InternVL-2.5~\cite{chen2024expanding}, Gemma3~\cite{team2025gemma3}, LLaVA-Onevision~\cite{li2024llava_ov}. 
To support retrieved ICL, we use a multimodal ensemble retriever combining textual and visual inputs. BERT-based text encoder~\cite{bge_embedding} encodes and retrieves text examples based on query outcomes, while a CLIP-based image encoder~\cite{radford2021clip} handles images. The ensemble merges similarity scores from both modalities with hyperparameters $\alpha$ assigned to $0.4$. For experiments, we created a database with 372 and 344 image-scenario pairs from MUN-vis and MUN-lang, which lack human label explanations and are not used for testing.
For baseline comparisons, we implement standard in-context learning (ICL) prompts where examples are randomly chosen from the MUN dataset, irrespective of their relevance to the query. 

\begin{table}[t]
\centering
\resizebox{\linewidth}{!}{
\begin{tabular}{lcccc}
\toprule
\textbf{Model} &
\textbf{LR} &
\textbf{LC} &
\textbf{LE} &
\textbf{CS} \\
\midrule
\textit{Gemma3}           & 3.16 (\small +0.12) & 3.59 (\small +0.15) & 4.12 (\small +0.14) & 3.59 (\small +0.14) \\
\textit{InternVL 2.5}     & 2.83 (\small +0.45) & 3.40 (\small +0.52) & 3.79 (\small +0.63) & 3.33 (\small +0.54) \\
\textit{LLaVA-OV}   & 3.21 (\small +0.07) & 3.77 (\small +0.07) & 4.13 (\small +0.04) & 3.84 (\small +0.11) \\
\textit{Phi 3.5v}         & 3.23 (\small +0.14) & 3.75 (\small +0.09) & 4.16 (\small +0.11) & 3.77 (\small +0.15) \\
\textit{Phi 4mm}           & 3.31 (\small +0.21) & 3.85 (\small +0.29) & 4.11 (\small +0.25) & 3.82 (\small +0.28) \\
\textit{Qwen2.5 VL}      & 3.29 (\small +0.08) & 3.87 (\small +0.13) & 4.25 (\small +0.07) & 3.89 (\small +0.12) \\
\textit{Qwen2 vl}          & 3.28 (\small +0.11) & 3.86 (\small +0.14) & 4.22 (\small +0.11) & 3.94 (\small +0.15) \\
\bottomrule
\end{tabular}
}
\caption{Effect of retrieval–based in-context selection on flask-based skill metrics(higher is better). LR stands for Logical Robustness, LC for Logical Correctness, LE for Logical Efficiency, and CS stands for Commonsense. Each cell shows the R-ICL score with the gain over the random baseline in parentheses.}
\label{tab:flask_skillset_eval}
\end{table}

\subsection{Experimental Setup}
\noindent \textbf{Varying the Number of In-Context Examples.}
To investigate how the number of in-context examples affects model performance, we vary the number of retrieved exemplars (from 1, 3, to 5) provided to the models.
This setup allows us to assess the scalability of the ICL approach and determine the optimal number of examples for effective reasoning.

\noindent \textbf{Retrieval-Based vs. Randomly Selected Shots.}
To evaluate the importance of retrieval quality, we compare our retrieval-based ICL with a baseline ICL approach that uses randomly selected examples from the MUN dataset. 

As for the metric, we adopted the Alpaca-Eval framework~\cite{alpaca_eval@alpaca_eval} to evaluate the quality of the generated explanations by comparing them against human and LLM-generated explanations. Specifically, we prompt GPT-4o to rank the explanations produced by different models against the Human+LLM explanations. The ranking assesses the coherence, relevance, and abductive reasoning quality of the model-generated explanations. 

\subsection{Results}
\label{ssec:main_results}

Table~\ref{tab:combined_results} summarizes the performance of the selected VLMs under various in-context example configurations and retrieval strategies. We report results for two tasks, MUN-vis and MUN-lang, to capture both visual and linguistic reasoning quality.

\noindent \textbf{RQ1. Effect of Shot Scaling.}  
All models generally improve as the number of in-context examples increases from zero to 1, 3, and 5 shots, especially when using R-ICL. 
Across the seven VLMs the median gain is \textbf{+6.1 pp} on \textit{vis} and \textbf{+7.2 pp} on \textit{lang}.
While MUN-lang also benefits from more examples and R-ICL, the improvement in MUN-vis is generally more pronounced, highlighting that visual reasoning gains more from the effective selection and increased number of in-context examples.

\noindent \textbf{RQ2. Retrieval \emph{vs} Random.}  
Comparing the two columns under each shot size in Table \ref{tab:combined_results} shows that R-ICL  beats random selection in \textbf{12 of 14} model-dataset combinations.  
The stronger gains on \textit{vis}(+0.040) over \textit{lang}(+0.023) confirm that supplying semantically aligned image exemplars is particularly helpful for visual reasoning.  
Notable examples include \emph{InternVL 2.5} (+13 pp at 1-shot) and \emph{Qwen2.5-VL} (+9.9 pp at 5-shot) on \textit{vis}; improvements on \textit{lang} are positive but smaller (e.g., \emph{Phi-4mm} +3.4 pp at 3-shot). 
Appendix \ref{sec:Qualitative results} offers a detailed qualitative analysis of these patterns and examples in the Qualitative Results section.

\subsection{Analysis of Model Response}

To rigorously quantify the contribution of R-ICL to the logical-reasoning capacities of VLMs, we conducted two complementary analyses: an automated evaluation and a human analysis.
\paragraph{Automatic evaluation}
In the automated, skill-based evaluation framework proposed by FLASK ~\cite{ye2023flask}, four complementary criteria are considered: Logical Robustness (LR), Logical Correctness (LC), Logical Efficiency (LE), and Commonsense Understanding (CS). Each scored on a 1-to-5 scale by GPT-4o using the rubric in the FLASK frameworks.
As summarized in Table \ref{tab:flask_skillset_eval}, retrieval-based in-context selection yields consistent improvements on every metric for every model. The largest absolute gains are observed for \emph{InternVL-2.5} (+0.63 LE; +0.54 CS), while even the smallest gains remain positive across all four skills. These findings indicate that supplying semantically relevant exemplars not only strengthens deductive reasoning but also enhances commonsense inference, further underscoring the role of context quality in in-context learning.
\begin{table}[t]
\centering
\scriptsize
\setlength{\tabcolsep}{6pt}
\begin{tabular}{lcc}
\toprule
\textbf{Setting} & \textbf{MUN-vis (\%)} & \textbf{MUN-lang (\%)} \\
\midrule
Zero-shot         & 16  & 36 \\
Random 5-shot     & 12  & 44 \\
Retrieval 5-shot  & 32  & 56 \\
\bottomrule
\end{tabular}
\vspace{4pt}
\caption{\textbf{Human preference win rates (\%).} Percentage of cases (out of 50 samples per modality) where human annotators preferred the model-generated explanation over the LLM+Human.}
\label{tab:human_win_rate}
\end{table}

\paragraph{Human evaluation}
Table~\ref{tab:human_win_rate} presents the results of a human evaluation comparing the reasoning quality of \emph{phi-4-mm} across different shot settings. For both the vision and language modalities, 50 samples were randomly selected, and human annotators were asked to compare the model’s responses, generated under zero-shot, random 5-shot, and retrieved 5-shot settings, against human-assisted explanations. For each sample, the annotators selected the response they judged to be more coherent and convincing. The reported values represent the winning ratio, i.e., the proportion of cases in which the model's output was preferred over the human-assisted explanation.  While some variance exists due to limited sample size, the overall trend, particularly in human evaluation, suggests that increasing the number of in-context examples, especially through retrieval, generally leads to improved reasoning performance.

\section{Conclusion}
We introduce the Multimodal UNcommonsense (MUN) dataset to evaluate how vision-language models handle atypical scenarios that challenge commonsense reasoning. Extensive experiments show that retrieved in-context learning (ICL) examples, rather than randomly chosen ones, enhance model performance. By bridging unexpected visual cues with logical explanations, we successfully guide models to produce more coherent, contextually aligned reasoning. 
This approach enables more adaptive and reliable multimodal AI systems that are better equipped to understand uncommon events, cultural nuances, and low-frequency phenomena in real-world settings.

\section{Acknowledgements}
This research was supported by Hyundai Motor Company and Kia Corporation. It was also funded by the Institute of Information \& Communications Technology Planning \& Evaluation (IITP) grant funded by the Korean government (MSIT) (No. RS-2020-II201361, Artificial Intelligence Graduate School Program, Yonsei University) and by the National Research Foundation of Korea (NRF) grants funded by the Korean government (MSIT) (Nos. RS-2024-00354218 and RS-2024-00353125). We also used AI assistants to refine the writing style and for preliminary coding assistance.

\section{Limitation}\label{sec:limitation}
While MUN provides a valuable benchmark for evaluating multimodal uncommonsense reasoning, it is not without shortcomings. First, while the dataset benefits from meticulous human curation that enhances per-sample quality, this comes at the cost of scale, potentially limiting its representation of the broader variability found in real-world scenarios and may not capture the full breadth of cultural, environmental, or domain-specific complexities. 

Second, our retrieval-based in-context learning approach, while effective, relies on the quality and diversity of available exemplars; overly domain-specific or homogeneous retrieval sets could limit the generalizability of results. 

Additionally, the current approach relies on post hoc evaluations with language models to assess explanation quality, which may introduce biases or yield incomplete metrics for reasoning capabilities. Subsequent efforts might also integrate multi-turn interactive reasoning processes, allowing models to clarify ambiguities before producing their final explanations. 
Advances in automated evaluation metrics could provide more objective assessments of abductive reasoning quality. 

Moreover, combining retrieval-based techniques with model fine-tuning or parameter-efficient adaptation strategies may yield more robust and domain-transferable reasoning systems. Ultimately, pursuing these directions can further strengthen the utility, fairness, and resilience of multimodal AI models in handling complex and atypical scenarios.

\section{Ethical Considerations}\label{ethical}

\textbf{Dataset Construction.} The dataset was constructed using images sourced from the web and carefully filtered to minimize inappropriate, sensitive content. All images were reviewed by annotators following a strict set of guidelines to ensure that the dataset does not propagate bias, stereotypes, or harmful cultural depictions.

\noindent
\textbf{Cultural and Contextual Reasoning.} The reasoning tasks presented in MUN encourage models to produce abductive explanations grounded in cultural and contextual knowledge. This raises the possibility that models might inadvertently generate content that reflects implicit biases or culturally insensitive narratives. We emphasize the importance of using diverse sets of evaluators and retrieval corpora to mitigate these risks and improve fairness and inclusivity. Researchers, developers, and users are encouraged to apply adversarial testing and ongoing monitoring to identify and address any unintended harm.

\textbf{Responsible Applications and Safeguards.} Lastly, while the improved reasoning capabilities we pursue may have beneficial real-world applications, from more accurate image analysis in healthcare to a better understanding of global cultural phenomena, they also open the door to more sophisticated image and text manipulation. It is crucial that developers implement robust guardrails, transparency measures, and user consent mechanisms to ensure that these advanced reasoning techniques serve the public interest responsibly, respecting privacy, cultural values, and intellectual property rights.

\bibliography{custom}

@inproceedings{li2022blip,
  title={Blip: Bootstrapping language-image pre-training for unified vision-language understanding and generation},
  author={Li, Junnan and Li, Dongxu and Xiong, Caiming and Hoi, Steven},
  booktitle={International conference on machine learning},
  pages={12888--12900},
  year={2022},
  organization={PMLR}
}

@article{alayrac2022flamingo,
  title={Flamingo: a visual language model for few-shot learning},
  author={Alayrac, Jean-Baptiste and Donahue, Jeff and Luc, Pauline and Miech, Antoine and Barr, Iain and Hasson, Yana and Lenc, Karel and Mensch, Arthur and Millican, Katherine and Reynolds, Malcolm and others},
  journal={Advances in neural information processing systems},
  volume={35},
  pages={23716--23736},
  year={2022}
}

@misc{bge_embedding,
      title={C-Pack: Packaged Resources To Advance General Chinese Embedding}, 
      author={Shitao Xiao and Zheng Liu and Peitian Zhang and Niklas Muennighoff},
      year={2023},
      eprint={2309.07597},
      archivePrefix={arXiv},
      primaryClass={cs.CL}
}

@inproceedings{radford2021clip,
  title={Learning transferable visual models from natural language supervision},
  author={Radford, Alec and Kim, Jong Wook and Hallacy, Chris and Ramesh, Aditya and Goh, Gabriel and Agarwal, Sandhini and Sastry, Girish and Askell, Amanda and Mishkin, Pamela and Clark, Jack and others},
  booktitle={International conference on machine learning},
  pages={8748--8763},
  year={2021},
  organization={PMLR}
}

@article{achiam2023gpt,
  title={Gpt-4 technical report},
  author={Achiam, Josh and Adler, Steven and Agarwal, Sandhini and Ahmad, Lama and Akkaya, Ilge and Aleman, Florencia Leoni and Almeida, Diogo and Altenschmidt, Janko and Altman, Sam and Anadkat, Shyamal and others},
  journal={arXiv preprint arXiv:2303.08774},
  year={2023}
}

@article{zhao2023uncommonsense,
  title={UNcommonsense Reasoning: Abductive Reasoning about Uncommon Situations},
  author={Zhao, Wenting and Chiu, Justin T and Hwang, Jena D and Brahman, Faeze and Hessel, Jack and Choudhury, Sanjiban and Choi, Yejin and Li, Xiang Lorraine and Suhr, Alane},
  journal={arXiv preprint arXiv:2311.08469},
  year={2023}
}

@article{lin2022unsupervised,
  title={Unsupervised cross-task generalization via retrieval augmentation},
  author={Lin, Bill Yuchen and Tan, Kangmin and Miller, Chris and Tian, Beiwen and Ren, Xiang},
  journal={Advances in Neural Information Processing Systems},
  volume={35},
  pages={22003--22017},
  year={2022}
}

@inproceedings{lin2023unlocking,
  title={The unlocking spell on base llms: Rethinking alignment via in-context learning},
  author={Lin, Bill Yuchen and Ravichander, Abhilasha and Lu, Ximing and Dziri, Nouha and Sclar, Melanie and Chandu, Khyathi and Bhagavatula, Chandra and Choi, Yejin},
  booktitle={The Twelfth International Conference on Learning Representations},
  year={2023}
}

@article{brown2020language,
  title={Language models are few-shot learners},
  author={Brown, Tom and Mann, Benjamin and Ryder, Nick and Subbiah, Melanie and Kaplan, Jared D and Dhariwal, Prafulla and Neelakantan, Arvind and Shyam, Pranav and Sastry, Girish and Askell, Amanda and others},
  journal={Advances in neural information processing systems},
  volume={33},
  pages={1877--1901},
  year={2020}
}

@inproceedings{chowdhery2022palm,
  title={PaLM: Scaling language modeling with Pathways},
  author={Chowdhery, Aakanksha and et al.},
  booktitle={arXiv preprint arXiv:2204.02311},
  year={2022}
}

@inproceedings{li2023blip2,
  title={BLIP-2: Bootstrapping Language-Image Pre-training with Frozen Image Encoders and Large Language Models},
  author={Li, Junnan and Li, Dongxu and Xie, Xiaohua and Yatskar, Mark and Hoi, Steven C.H.},
  booktitle={ICML},
  year={2023}
}

@article{thoppilan2022lamda,
  title={Lamda: Language models for dialog applications},
  author={Thoppilan, Romal and De Freitas, Daniel and Hall, Jamie and Shazeer, Noam and Kulshreshtha, Apoorv and Cheng, Heng-Tze and Jin, Alicia and Bos, Taylor and Baker, Leslie and Du, Yu and others},
  journal={arXiv preprint arXiv:2201.08239},
  year={2022}
}

@article{wei2022chain,
  title={Chain-of-thought prompting elicits reasoning in large language models},
  author={Wei, Jason and Wang, Xuezhi and Schuurmans, Dale and Bosma, Maarten and Xia, Fei and Chi, Ed and Le, Quoc V and Zhou, Denny and others},
  journal={Advances in neural information processing systems},
  volume={35},
  pages={24824--24837},
  year={2022}
}

@article{zhou2022least,
  title={Least-to-most prompting enables complex reasoning in large language models},
  author={Zhou, Denny and Sch{\"a}rli, Nathanael and Hou, Le and Wei, Jason and Scales, Nathan and Wang, Xuezhi and Schuurmans, Dale and Cui, Claire and Bousquet, Olivier and Le, Quoc and others},
  journal={arXiv preprint arXiv:2205.10625},
  year={2022}
}

@article{zhu2020dark,
  title={Dark, beyond deep: A paradigm shift to cognitive ai with humanlike common sense},
  author={Zhu, Yixin and Gao, Tao and Fan, Lifeng and Huang, Siyuan and Edmonds, Mark and Liu, Hangxin and Gao, Feng and Zhang, Chi and Qi, Siyuan and Wu, Ying Nian and others},
  journal={Engineering},
  volume={6},
  number={3},
  pages={310--345},
  year={2020},
  publisher={Elsevier}
}

@article{shao2024cpt,
  title={Cpt: A pre-trained unbalanced transformer for both chinese language understanding and generation},
  author={Shao, Yunfan and Geng, Zhichao and Liu, Yitao and Dai, Junqi and Yan, Hang and Yang, Fei and Li, Zhe and Bao, Hujun and Qiu, Xipeng},
  journal={Science China Information Sciences},
  volume={67},
  number={5},
  pages={152102},
  year={2024},
  publisher={Springer}
}

@inproceedings{bhagavatula2019abductive,
  title={Abductive Commonsense Reasoning},
  author={Bhagavatula, Chandra and Le Bras, Ronan and Malaviya, Chaitanya and Sakaguchi, Keisuke and Holtzman, Ari and Rashkin, Hannah and Downey, Doug and Yih, Wen-tau and Choi, Yejin},
  booktitle={International Conference on Learning Representations},
  url={https://openreview.net/forum?id=Byg1v1HKDB},
  year={2020},
}

@inproceedings{hessel2022abduction,
  title={The abduction of sherlock holmes: A dataset for visual abductive reasoning},
  author={Hessel, Jack and Hwang, Jena D and Park, Jae Sung and Zellers, Rowan and Bhagavatula, Chandra and Rohrbach, Anna and Saenko, Kate and Choi, Yejin},
  booktitle={European Conference on Computer Vision},
  pages={558--575},
  year={2022},
  organization={Springer}
}

@article{ventura2024nl,
  title={NL-Eye: Abductive NLI for Images},
  author={Ventura, Mor and Toker, Michael and Calderon, Nitay and Gekhman, Zorik and Bitton, Yonatan and Reichart, Roi},
  journal={arXiv preprint arXiv:2410.02613},
  year={2024}
}

@inproceedings{han-etal-2023-reading,
    title = "Reading Books is Great, But Not if You Are Driving! Visually Grounded Reasoning about Defeasible Commonsense Norms",
    author = "Han, Seungju  and
      Kim, Junhyeok  and
      Hessel, Jack  and
      Jiang, Liwei  and
      Chung, Jiwan  and
      Son, Yejin  and
      Choi, Yejin  and
      Yu, Youngjae",
    editor = "Bouamor, Houda  and
      Pino, Juan  and
      Bali, Kalika",
    booktitle = "Proceedings of the 2023 Conference on Empirical Methods in Natural Language Processing",
    month = dec,
    year = "2023",
    address = "Singapore",
    publisher = "Association for Computational Linguistics",
    url = "https://aclanthology.org/2023.emnlp-main.57",
    doi = "10.18653/v1/2023.emnlp-main.57",
    pages = "894--914",
}

@article{li2021align,
  title={Align before fuse: Vision and language representation learning with momentum distillation},
  author={Li, Junnan and Selvaraju, Ramprasaath and Gotmare, Akhilesh and Joty, Shafiq and Xiong, Caiming and Hoi, Steven Chu Hong},
  journal={Advances in neural information processing systems},
  volume={34},
  pages={9694--9705},
  year={2021}
}

@misc{liu2023llava,
  title     = {{LLaVA}: Large Language and Vision Assistant},
  author    = {Liu, Haotian and Li, Chunyuan and Wu, Qingyang and Lee, Yong Jae},
  year      = {2023},
  eprint    = {2302.06675},
  archivePrefix = {arXiv},
  primaryClass = {cs.CL}
}

@misc{schuhmann2022laion,
  title     = {{LAION-5B}: An open large-scale dataset for training next generation image-text models},
  author    = {Schuhmann, Christoph and Beaumont, Lyonel and Vencu, Romain and Gordon, Cade W. and Wightman, Ross and Cherti, Mehdi and Coombes, Theo and Katta, Kartikay and Mullis, Chris and Kaczmarczyk, Roman and others},
  year      = {2022},
  eprint    = {2210.08402},
  archivePrefix = {arXiv},
  primaryClass = {cs.CV}
}

@inproceedings{Stiennon2020,
  title={Learning to summarize with human feedback},
  author={Stiennon, Nisan and Ouyang, Long and Wu, Jeffrey and Ziegler, Daniel M and Lowe, Ryan and Voss, Casey and Radford, Alec and Amodei, Dario and Sutskever, Ilya},
  booktitle={Advances in Neural Information Processing Systems},
  year={2020}
}

@article{maaten2008visualizing,
  title={Visualizing data using t-SNE},
  author={Maaten, Laurens van der and Hinton, Geoffrey},
  journal={Journal of machine learning research},
  volume={9},
  number={Nov},
  pages={2579--2605},
  year={2008}
}

@inproceedings{Schwenk2022AOKVQA,
  title={A-OKVQA: A Benchmark for Visual Question Answering using World Knowledge},
  author={Dustin Schwenk and Apoorv Khandelwal and Christopher Clark and Kenneth Marino and Roozbeh Mottaghi},
  booktitle={European Conference on Computer Vision},
  year={2022},
  url={https://api.semanticscholar.org/CorpusID:249375629}
}

@article{wang2024qwen2,
  title={Qwen2-vl: Enhancing vision-language model's perception of the world at any resolution},
  author={Wang, Peng and Bai, Shuai and Tan, Sinan and Wang, Shijie and Fan, Zhihao and Bai, Jinze and Chen, Keqin and Liu, Xuejing and Wang, Jialin and Ge, Wenbin and others},
  journal={arXiv preprint arXiv:2409.12191},
  year={2024}
}

@article{bai2025qwen2_5vl,
  title={Qwen2. 5-vl technical report},
  author={Bai, Shuai and Chen, Keqin and Liu, Xuejing and Wang, Jialin and Ge, Wenbin and Song, Sibo and Dang, Kai and Wang, Peng and Wang, Shijie and Tang, Jun and others},
  journal={arXiv preprint arXiv:2502.13923},
  year={2025}
}

@article{abdin2024phi,
  title={Phi-3 technical report: A highly capable language model locally on your phone},
  author={Abdin, Marah and Aneja, Jyoti and Awadalla, Hany and Awadallah, Ahmed and Awan, Ammar Ahmad and Bach, Nguyen and Bahree, Amit and Bakhtiari, Arash and Bao, Jianmin and Behl, Harkirat and others},
  journal={arXiv preprint arXiv:2404.14219},
  year={2024}
}

@article{abouelenin2025phi4,
  title={Phi-4-mini technical report: Compact yet powerful multimodal language models via mixture-of-loras},
  author={Abouelenin, Abdelrahman and Ashfaq, Atabak and Atkinson, Adam and Awadalla, Hany and Bach, Nguyen and Bao, Jianmin and Benhaim, Alon and Cai, Martin and Chaudhary, Vishrav and Chen, Congcong and others},
  journal={arXiv preprint arXiv:2503.01743},
  year={2025}
}

@article{team2025gemma3,
  title={Gemma 3 technical report},
  author={Team, Gemma and Kamath, Aishwarya and Ferret, Johan and Pathak, Shreya and Vieillard, Nino and Merhej, Ramona and Perrin, Sarah and Matejovicova, Tatiana and Ram{\'e}, Alexandre and Rivi{\`e}re, Morgane and others},
  journal={arXiv preprint arXiv:2503.19786},
  year={2025}
}

@article{li2024llava_ov,
  	title={LLaVA-OneVision: Easy Visual Task Transfer},
  	author={Li, Bo and Zhang, Yuanhan and Guo, Dong and Zhang, Renrui and Li, Feng and Zhang, Hao and Zhang, Kaichen and Li, Yanwei and Liu, Ziwei and Li, Chunyuan},
  	journal={arXiv preprint arXiv:2408.03326},
  	year={2024}
}

@article{chen2024expanding,
  title={Expanding Performance Boundaries of Open-Source Multimodal Models with Model, Data, and Test-Time Scaling},
  author={Chen, Zhe and Wang, Weiyun and Cao, Yue and Liu, Yangzhou and Gao, Zhangwei and Cui, Erfei and Zhu, Jinguo and Ye, Shenglong and Tian, Hao and Liu, Zhaoyang and others},
  journal={arXiv preprint arXiv:2412.05271},
  year={2024}
}

@misc{alpaca_eval@alpaca_eval,
  author = {Xuechen Li and Tianyi Zhang and Yann Dubois and Rohan Taori and Ishaan Gulrajani and Carlos Guestrin and Percy Liang and Tatsunori B. Hashimoto },
  title = {AlpacaEval: An Automatic Evaluator of Instruction-following Models},
  year = {2023},
  month = {5},
  publisher = {GitHub},
  journal = {GitHub repository},
  howpublished = {\url{https://github.com/tatsu-lab/alpaca_eval}}
}

@article{wang2023commonsensevis,
  title={CommonsenseVIS: Visualizing and understanding commonsense reasoning capabilities of natural language models},
  author={Wang, Xingbo and Huang, Renfei and Jin, Zhihua and Fang, Tianqing and Qu, Huamin},
  journal={IEEE Transactions on Visualization and Computer Graphics},
  year={2023},
  publisher={IEEE}
}

@article{tao2024cultural,
  title={Cultural bias and cultural alignment of large language models},
  author={Tao, Yan and Viberg, Olga and Baker, Ryan S and Kizilcec, Ren{\'e} F},
  journal={PNAS nexus},
  volume={3},
  number={9},
  pages={pgae346},
  year={2024},
  publisher={Oxford University Press US}
}

@article{bosselut2019comet,
  title={COMET: Commonsense transformers for automatic knowledge graph construction},
  author={Bosselut, Antoine and Rashkin, Hannah and Sap, Maarten and Malaviya, Chaitanya and Celikyilmaz, Asli and Choi, Yejin},
  journal={arXiv preprint arXiv:1906.05317},
  year={2019}
}

@article{ye2023flask,
  title={Flask: Fine-grained language model evaluation based on alignment skill sets},
  author={Ye, Seonghyeon and Kim, Doyoung and Kim, Sungdong and Hwang, Hyeonbin and Kim, Seungone and Jo, Yongrae and Thorne, James and Kim, Juho and Seo, Minjoon},
  journal={arXiv preprint arXiv:2307.10928},
  year={2023}
}

@article{raffel2020exploring,
  title = {Exploring the Limits of Transfer Learning with a Unified Text-to-Text Transformer},
  author = {Raffel, Colin and Shazeer, Noam and Roberts, Adam and Lee, Katherine and Narang, Sharan and Matena, Niki and Parmar, Niki and Liu, Yanqi and Jolicoeur-Martineau, Alex},
  journal = {Journal of Machine Learning Research},
  volume = {21},
  year = {2020},
  pages = {1--67}
}

@inproceedings{hendricks2016generating,
  title = {Generating Visual Explanations},
  author = {Hendricks, Lisa Anne and Venugopalan, Subhashini and Rohrbach, Marcus and Mooney, Raymond and Saenko, Kate and Darrell, Trevor},
  booktitle = {European Conference on Computer Vision (ECCV)},
  pages = {3--19},
  year = {2016}
}

@inproceedings{agrawal2017cvt,
  title = {Don't Just Assume; Look and Answer: Overcoming Priors for Visual Question Answering},
  author = {Agrawal, Aishwarya and Batra, Dhruv and Parikh, Devi},
  booktitle = {Conference on Computer Vision and Pattern Recognition (CVPR)},
  year = {2017}
}

@article{li2019visualbert,
  title = {{VisualBERT}: A Simple and Performant Baseline for Vision and Language},
  author = {Li, Liunian Harold and Yatskar, Mark and Yin, Da and Hsieh, Cho-Jui and Chang, Kai-Wei and Chi, Zhenyu},
  journal = {arXiv preprint arXiv:1908.03557},
  year = {2019}
}

@inproceedings{gao-etal-2021-simcse,
    title = "{S}im{CSE}: Simple Contrastive Learning of Sentence Embeddings",
    author = "Gao, Tianyu  and
      Yao, Xingcheng  and
      Chen, Danqi",
    editor = "Moens, Marie-Francine  and
      Huang, Xuanjing  and
      Specia, Lucia  and
      Yih, Scott Wen-tau",
    booktitle = "Proceedings of the 2021 Conference on Empirical Methods in Natural Language Processing",
    month = nov,
    year = "2021",
    address = "Online and Punta Cana, Dominican Republic",
    publisher = "Association for Computational Linguistics",
    url = "https://aclanthology.org/2021.emnlp-main.552",
    doi = "10.18653/v1/2021.emnlp-main.552",
    pages = "6894--6910",
}

@inproceedings{kwon2023efficient,
  title={Efficient memory management for large language model serving with pagedattention},
  author={Kwon, Woosuk and Li, Zhuohan and Zhuang, Siyuan and Sheng, Ying and Zheng, Lianmin and Yu, Cody Hao and Gonzalez, Joseph and Zhang, Hao and Stoica, Ion},
  booktitle={Proceedings of the 29th Symposium on Operating Systems Principles},
  pages={611--626},
  year={2023}
}

@article{douze2024faiss,
  title={The faiss library},
  author={Douze, Matthijs and Guzhva, Alexandr and Deng, Chengqi and Johnson, Jeff and Szilvasy, Gergely and Mazar{\'e}, Pierre-Emmanuel and Lomeli, Maria and Hosseini, Lucas and J{\'e}gou, Herv{\'e}},
  journal={arXiv preprint arXiv:2401.08281},
  year={2024}
}

@inproceedings{liu-etal-2023-g,
    title = "{G}-Eval: {NLG} Evaluation using Gpt-4 with Better Human Alignment",
    author = "Liu, Yang  and
      Iter, Dan  and
      Xu, Yichong  and
      Wang, Shuohang  and
      Xu, Ruochen  and
      Zhu, Chenguang",
    editor = "Bouamor, Houda  and
      Pino, Juan  and
      Bali, Kalika",
    booktitle = "Proceedings of the 2023 Conference on Empirical Methods in Natural Language Processing",
    month = dec,
    year = "2023",
    address = "Singapore",
    publisher = "Association for Computational Linguistics",
    url = "https://aclanthology.org/2023.emnlp-main.153/",
    doi = "10.18653/v1/2023.emnlp-main.153",
    pages = "2511--2522",
   
}

\newpage
\appendix
\onecolumn
\section{Experiment Details and Hyperparameter}
\label{sec:hparams}
Table \ref{tab:hparams} shows the hyperparameters of the models we used in the experiments and the exact model checkpoints used in the experiments are reported in Table \ref{tab:checkpoints}.
All experiments, except those involving GPT-4o, were conducted using two NVIDIA A6000 GPUs.

\begin{table}[H]
\centering
\resizebox{0.7\linewidth}{!}{
\begin{tabular}{@{}l|l@{}}
\toprule
\textbf{Hyperparameter} & \textbf{Configuration} \\ \midrule
Text emb. model & \texttt{BAAI/bge-large-en} \\
Image emb. model & \texttt{clip-vit-base-patch16} \\
Image resolution & 512×512 \\
Ensemble ratio $\alpha$ & 0.4 \\
Retrieval lib. & \texttt{langchain} (\url{https://python.langchain.com/docs/introduction/}) \\
Vector DB lib. & FAISS~\cite{douze2024faiss} \\
VLLM lib. & VLLM~\cite{kwon2023efficient} \\
\bottomrule
\end{tabular}
}
\caption{Hyperparameter configurations for the main experiment.}
\label{tab:hparams}
\end{table}

\begin{table}[h]
  \centering
  \setlength{\tabcolsep}{6pt}
  \resizebox{0.5\linewidth}{!}{
  \begin{tabular}{ll}
    \toprule
    \multicolumn{2}{l}{\textbf{Open-source Models}} \\
    \midrule
    \textit{Gemma 3}           & \texttt{google/gemma-3-4b-it} \\
    \textit{InternVL 2.5}      & \texttt{OpenGVLab/InternVL2\_5-8B} \\
    \textit{LLaVA-OneVision}   & \texttt{llava-hf/llava-onevision-qwen2-7b-ov-hf} \\
    \textit{Phi 3.5-Vision}    & \texttt{microsoft/Phi-3.5-vision-instruct} \\
    \textit{Phi 4-Multimodal}  & \texttt{microsoft/Phi-4-multimodal-instruct} \\
    \textit{Qwen 2.5-VL}       & \texttt{Qwen/Qwen2.5-VL-7B-Instruct} \\
    \textit{Qwen 2-VL}         & \texttt{Qwen/Qwen2-VL-7B-Instruct} \\
    \midrule
    \multicolumn{2}{l}{\textbf{Closed-source Models}} \\
    \midrule
    \textit{GPT-4o}            & \texttt{gpt-4o-2024-11-20 (via OpenAI API)} \\
    \bottomrule
  \end{tabular}
  }
  \caption{
    Model checkpoints used in our experiments. Open-source models were accessed via Hugging Face, and the closed-source model (GPT-4o) was accessed via the OpenAI API.
  }
  \label{tab:checkpoints}
\end{table}

\section{Bi-Encoder Retrieval Mechanism}
To retrieve relevant in-context examples for uncommonsense reasoning, we use a bi-encoder retrieval strategy that computes and fuses modality-specific similarity scores. First, we embed the (image, text) pairs stored in the dataset $D_{(i,t)}$ using a CLIP-style image encoder $E_I$ and a BERT-based text encoder $E_T$, respectively. Given a user query $q = (q_i, q_t)$, we compute cosine similarities between the query vectors $(v_{q_i}, v_{q_t})$ and stored vectors $(v_i, v_t)$. The final similarity score is obtained by weighting the image and text similarities using a tunable coefficient $\alpha$ that controls the relative contribution of each modality.

\label{sec:retrieval_algo}
\begin{algorithm}[H]
\begin{algorithmic}[1]
\small
\STATE \textbf{Input:} $q = (q_i, q_t)$; number of retrievals $k$, weight ratio $\alpha$
\STATE \textbf{Output:} list of top $k$ retrieved $(d_i, d_t)$ pairs

\STATE Vector database $D_{(i, t)}$ containing (image, text) pairs
\STATE Image encoder $E_i$ and text encoder $E_t$
\STATE Convert query to vectors: $v_q = (v_{q_i}, v_{q_t}) = (E_i(q_i), E_t(q_t))$
\STATE Initialize \texttt{Results} $\gets []$, \texttt{Indices} $\gets []$

\FOR{each $(v_i, v_t)$ with index $j$ in $D_{(i, t)}$}
    \STATE Compute similarity: 
    \STATE $s = \alpha \cdot \text{cos\_sim}(v_{q_i}, v_i) + (1-\alpha) \cdot \text{cos\_sim}(v_{q_t}, v_t)$
    \STATE Append $s$ to \texttt{Results}
    \STATE Append $j$ to \texttt{Indices}
\ENDFOR

\STATE Combine \texttt{Results} and \texttt{Indices} into pairs and sort by similarity in descending order
\STATE Select top $k$ indices as \texttt{TopKIndices}
\STATE Initialize \texttt{TopK} $\gets []$
\FOR{each j in TopKIndices}
    \STATE Append $(d_{i_j}, d_{t_j})$ to \texttt{TopK}
\ENDFOR
\STATE \textbf{return} \texttt{TopK}
\end{algorithmic}
\caption{Ensemble Retrieval Method (Bi-encoder). This computes cosine similarities in visual and textual embedding spaces, fuses them by $\alpha$, and returns the top $k$ matches.}
\label{alg:verification}
\end{algorithm}

\section{Qualitative results}\label{sec:Qualitative results}

\begin{figure*}[!ht]
\centering
\includegraphics[trim={0 0cm 0 0},width=\linewidth]{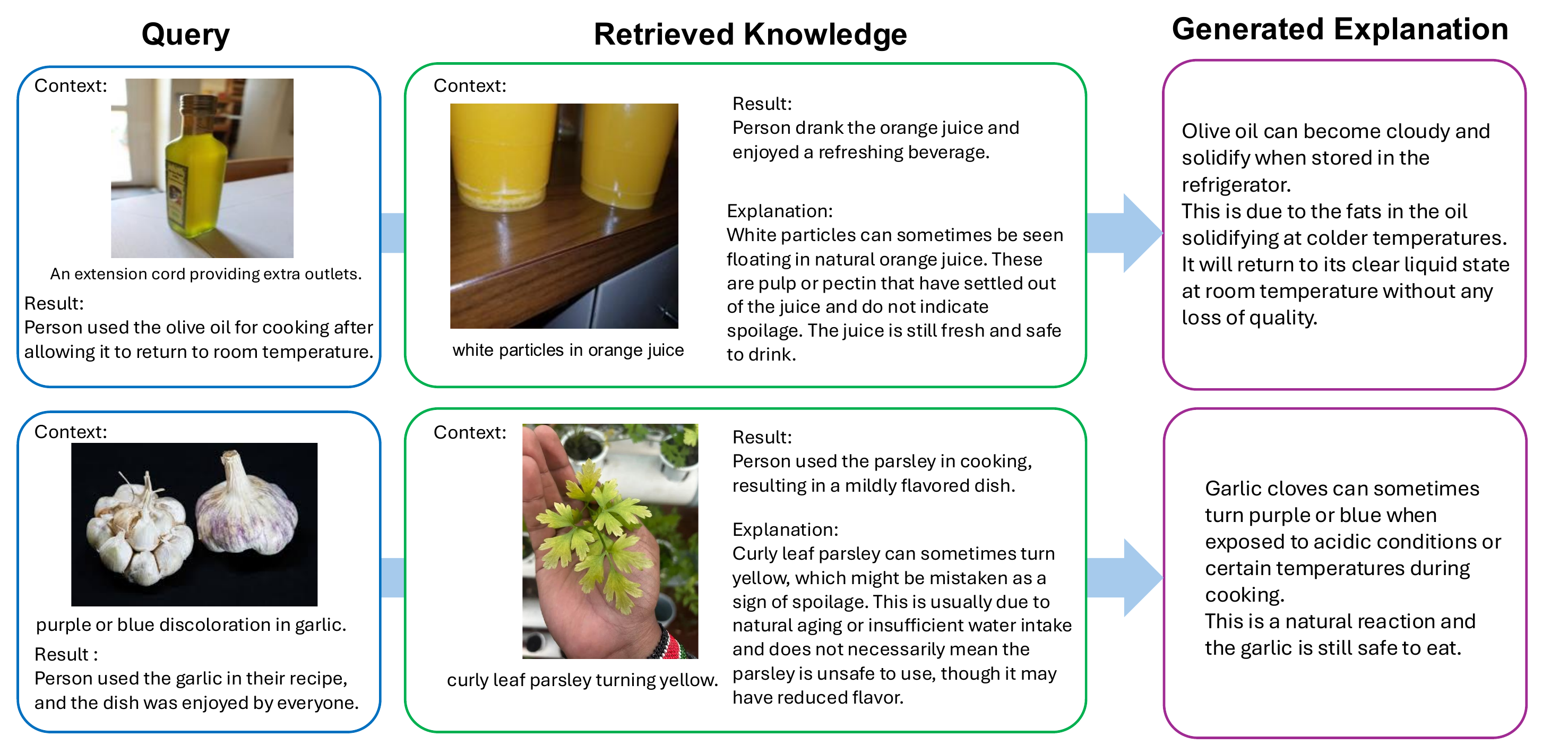}
\caption{MUN-vis qualitative results.}
\label{fig:QR_vis}
\end{figure*}

Figure \ref{fig:QR_vis} illustrates the model’s capacity to retrieve contextual knowledge and produce precise, explanatory answers in MUN-vis. For the first row, when queried about the haze that develops in refrigerated olive oil, the model draws an analogy to the white flecks that appear in orange juice. In both cases, low temperatures cause constituents to congeal and aggregate: fats solidify in olive oil, while pulp- and pectin-rich particles clump together in orange juice. Once the liquids return to room temperature, they clarify, showing that neither product’s quality is compromised. This example demonstrates how the system enhances its explanatory power by juxtaposing uncommon yet analogous phenomena across different contexts.  

\begin{figure*}[!ht]
\centering
\includegraphics[trim={0 0cm 0 0},width=\linewidth]{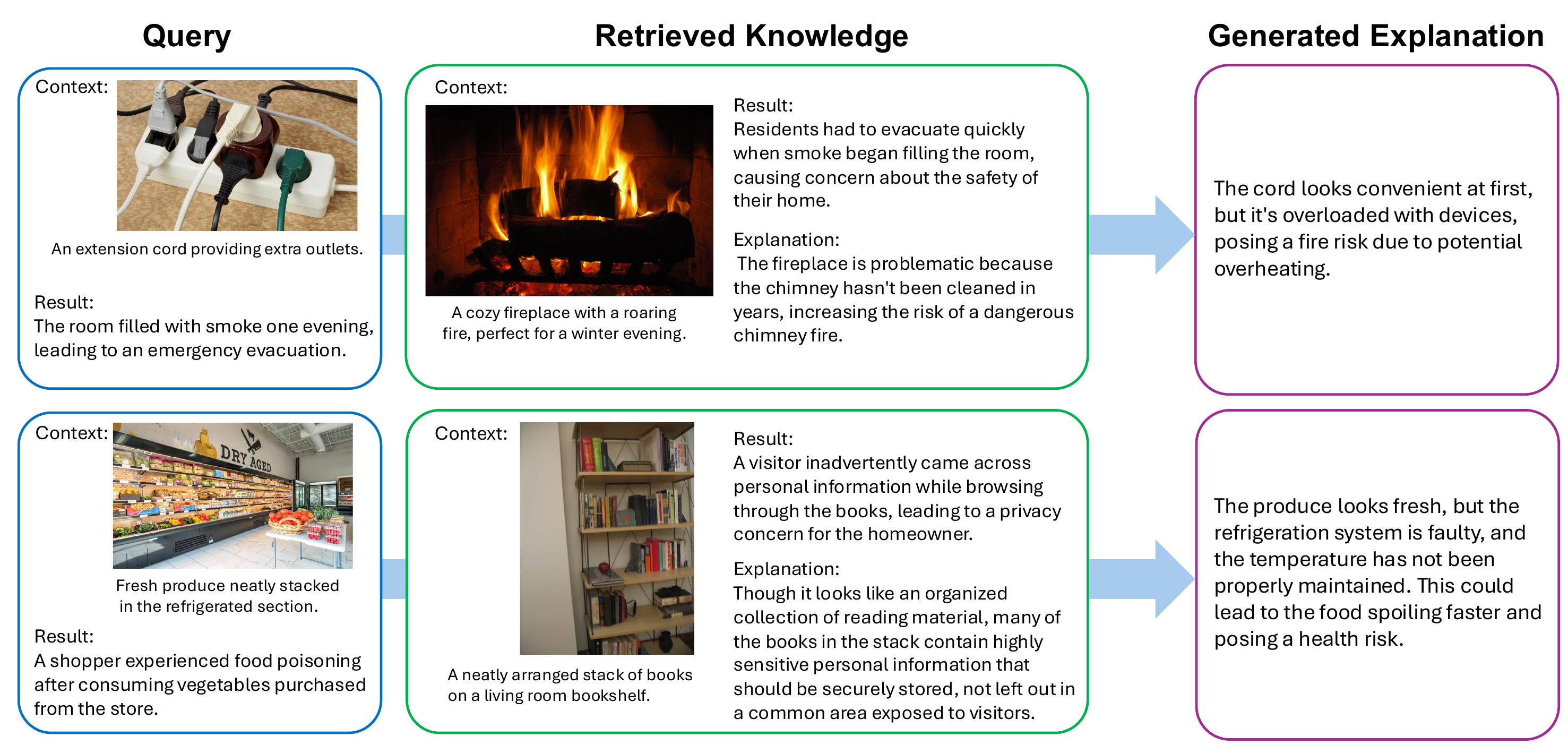}
\caption{ MUN-lang qualitative results. }
\label{fig:QR_lang}
\end{figure*}

Figure \ref{fig:QR_lang} highlights the model’s ability to retrieve contextual knowledge and generate precise, explanatory responses in MUN-lang. For the first row, while the power strip appears normal at first glance, the outcome of `smoke and emergency evacuation' necessitates the model to abductively infer an `inherent risk of overheating' within the power strip. To facilitate this inference, the retrieved few-shot examples include scenarios such as `residents evacuating due to smoke from a cozy fireplace'. Despite depicting different visual subjects, it induces a common causal pattern: `an ostensibly normal object contains a hidden fire hazard'. This connects to everyday experience-based abductive reasoning that `appliances generating heat pose a fire hazard', helping the model to infer potential dangers beyond what is visually apparent.

As the qualitative examples make clear, MUN-vis and MUN-lang probe two orthogonal yet complementary facets of uncommonsense reasoning. In MUN-vis, an uncommon visual cue must be normalized via specific commonsense knowledge (e.g., ``cloudy refrigerated olive oil'' $\leftrightarrow$ ``white flecks in chilled orange juice''
), whereas MUN-lang inverts the challenge: a common visual scene masks an anomalous outcome, demanding abductive reconstruction of hidden risks (e.g., ``benign-looking power strip'' $\rightarrow$ ``concealed fire hazard'' $\rightarrow$ ``evacuation''). Together, these tracks enforce a balanced assessment of a model’s ability to anchor striking images to everyday facts and infer unseen causal mechanisms behind unexpected events. 
By integrating both dimensions into a single benchmark and leveraging MER’s targeted retrieval of concrete analogues or abstract causal templates, MUN provides a comprehensive framework for evaluating models, spanning from concrete commonsense grounding to abstract causal inference.

\section{Comparison of Human Agreement on Explanations}
\label{sec:human_agreement_comparison}
We conduct a human evaluation comparing human-written explanations against those generated by two types of models: \textsc{LLM} (directly generated by the model) and \textsc{HLLM} (either model-generated or model-augmented based on human-written content).
As summarized in Table~\ref{tab:human_agreement_comparison}, there is no significant decline in the perceived quality of responses generated by the LLM. 
Specifically, 70.8\% of LLM-generated explanations achieved higher than moderate agreement with human-written explanations, while 71.6\% of HLLM explanations reached this level of agreement. 
These results indicate that model-generated and model-augmented explanations can closely match human-written ones in terms of response quality.

\section{Evaluation of GPT4o on MUN dataset}
\begin{table}[!t]
\centering
\resizebox{0.65\linewidth}{!}{
\begin{tabular}{p{1.8cm} | c c | c c | c c }
\toprule
\multirow{2}{*}{\textbf{Dataset}} &
\multicolumn{2}{c|}{\textbf{1-shot}} &
\multicolumn{2}{c|}{\textbf{3-shot}} &
\multicolumn{2}{c}{\textbf{5-shot}} \\
\cmidrule(lr){2-3}\cmidrule(lr){4-5}\cmidrule(lr){6-7}
& Rand. & R-ICL & 
           Rand. & R-ICL & 
           Rand. & R-ICL \\
\midrule
MUN vis & 0.572 & 0.597 & 0.604 & 0.610 & 0.673 & 0.704 \\
MUN lang & 0.678 & 0.636 & 0.671 & 0.664 & 0.650 & 0.692 \\
\bottomrule
\end{tabular}}
\vspace{-2mm}
\caption{Evaluation of GPT4o on different shot settings, measured by winning ratio against human-assisted explanations(higher is better). "Random" indicates randomly chosen examples, and "R-ICL" indicates retrieved examples for in-context learning. Model outputs were compared with Human+LLM explanations, judged using LLM. }
\label{tab:gpt4o_results}
\end{table}
We have conducted GPT-4o's performance on our dataset with a similar setup as the sec~\ref{sec:experiments}, which shows strong performance across both mun-vis and mun-lang, with generally similar performance improvement trends with open-source models. However, we excluded GPT-4o from our initial experiments due to the well-documented "self-preference bias" where LLMs tend to favor their own generated answers and attach our results in the appendix.

\section{Evaluation of haperparameter alpha}
\begin{table}[h]
\centering
\resizebox{0.5\textwidth}{!}{%
\begin{tabular}{c|ccccc}
\toprule
\textbf{$\alpha$}   & 0.3   & 0.4   & 0.5   & 0.6   & 0.7   \\ \midrule
\textbf{Winrate}    & 0.572 & 0.618 & 0.611 & 0.611 & 0.603 \\ \bottomrule
\end{tabular}
}
\caption{Ablation study on hyperparameter alpha on MUN vis with  Phi4-mm model.}
\label{tab:abl_alpha_exp}
\end{table}

Table~\ref{tab:abl_alpha_exp} shows the effects of different hyperparameter $\alpha$ on performance on the MER on a MUN-vis subset with Phi4-mm model. Based on findings in Table~\ref{tab:abl_alpha_exp}, we have used an $\alpha$ value of 0.4 during the main experiments.

\section{Evaluation with an Open-Source Judge}
\begin{table}[!t]
\centering
\resizebox{0.7\linewidth}{!}{
\begin{tabular}{p{0.85cm} l | c | c c | c c | c c }
\toprule
\multirow{2}{*}{\textbf{Dataset}} & \multirow{2}{*}{\textbf{~~~Model}} &
\multirow{2}{*}{\textbf{0-shot}} &
\multicolumn{2}{c|}{\textbf{1-shot}} &
\multicolumn{2}{c|}{\textbf{3-shot}} &
\multicolumn{2}{c}{\textbf{5-shot}} \\
\cmidrule(lr){4-5}\cmidrule(lr){6-7}\cmidrule(lr){8-9}
& &  & Rand. & R-ICL & Rand. & R-ICL & Rand. & R-ICL \\
\midrule
\multirow{4}{*}{\makecell[c]{\textbf{MUN}\\\textbf{lang}}}
& \textit{Gemma}3       & 0.259 & 0.364 & \textbf{0.399} & 0.294 & 0.301 & 0.259 & 0.287 \\
& \textit{Phi}3.5\textit{v}   & 0.273 & 0.329 & \textbf{0.336} & 0.287 & 0.329 & 0.315 & 0.287 \\
& \textit{Phi}4\textit{mm}    & 0.371 & \textbf{0.497} & 0.448 & 0.455 & 0.427 & 0.448 & \textbf{0.497} \\
& \textit{Qwen2.5}\textit{VL} & 0.364 & 0.322 & 0.357 & 0.287 & \textbf{0.378} & 0.273 & 0.322 \\
\midrule
\multirow{4}{*}{\makecell[c]{\textbf{MUN}\\\textbf{vis}}}
& \textit{Gemma}3       & 0.333 & 0.233 & 0.289 & 0.226 & \textbf{0.365} & 0.233 & 0.314 \\
& \textit{Phi}3.5\textit{v}   & 0.075 & 0.126 & 0.176 & 0.113 & \textbf{0.201} & 0.170 & 0.157 \\
& \textit{Phi}4\textit{mm}    & 0.195 & 0.239 & 0.214 & 0.132 & \textbf{0.277} & 0.101 & 0.170 \\
& \textit{Qwen2.5}\textit{VL} & 0.170 & 0.164 & \textbf{0.358} & 0.176 & 0.327 & 0.208 & 0.302 \\
\bottomrule
\end{tabular}}
\vspace{-2mm}
\caption{Comparison of models in different shot settings, measured by winning ratio against human-assisted explanations, judged by opensource LLM. "Random" indicates randomly chosen examples, and "R-ICL" indicates retrieved examples for in-context learning. Model outputs were compared with Human+LLM explanations, judged using opensource LLM(Llama-4-Scout).}
\label{tab:main_opensource_judge}
\end{table}
 To verify that the performance benefits of our R-ICL method are robust and not dependent on a single proprietary evaluator, we have evaluated with the state-of-the-art open-source Llama-4-Scout model as a judge model for comparison between model outputs and Human+LLM explanations. As tab~\ref{tab:main_opensource_judge} confirms that the central trend observed in the main experiments holds. While absolute win rates differ due to the new evaluator's distinct preferences, our Retrieval-Augmented In-Context Learning (R-ICL) consistently outperforms or remains highly competitive with zero-shot and random few-shot baselines across most models and settings (achievements highlighted in bold).

\section{Evaluation of MER on other opensource benchmarks}
\begin{table}[h]
\centering
\resizebox{0.3\textwidth}{!}{%
\begin{tabular}{c|c}
\toprule
Sampleing Mode & Accuracy \\ \midrule
Zero shot & 0.818 \\
Random 1 shot & 0.832 \\
R-ICL 1 shot & \textbf{0.842} \\
\bottomrule
\end{tabular}
}
\caption{Ablation study on MER method on A-OKVQA datasets with Qwen-2.5-VL.}
\label{tab:okvqa}
\end{table}
To provide empirical evidence for the generalizability of our MER framework, we conducted a preliminary experiment on the A-OKVQA benchmark~\cite{Schwenk2022AOKVQA}. We tested the accuracy of Qwen-2.5-VL on 500 randomly selected multiple-choice questions from the validation set with 5000 samples from the training set acting as ICL context. Table~\ref{tab:okvqa} demonstrates that R-ICL improves accuracy over both zero-shot and random-shot baselines. The performance gain on A-OKVQA, a task requiring both visual understanding and external knowledge, strongly suggests that MER's ability to retrieve relevant context is a generalizable principle.

\begin{table}[ht]
\centering
{\footnotesize
\begin{tabular}{c|rr|rr}
\toprule
\textbf{Level} & \multicolumn{2}{c|}{\textbf{LLM}} & \multicolumn{2}{c}{\textbf{HLLM}} \\
              & \textbf{Cnt} & \textbf{\%} & \textbf{Cnt} & \textbf{\%} \\
\midrule
1 & 136 & 13.6 & 123 & 12.3 \\
2 & 162 & 16.2 & 161 & 16.1 \\
3 & 194 & 19.4 & 204 & 20.4 \\
4 & 221 & 22.1 & 255 & 25.5 \\
5 & 287 & 28.7 & 257 & 25.7 \\
\midrule
\textbf{Avg.} & \multicolumn{2}{c|}{\textbf{3.36}} & \multicolumn{2}{c}{\textbf{3.36}} \\
\bottomrule
\end{tabular}
}
\caption{Distribution of human agreement levels (out of 1000 samples each) for LLM vs. Human and HLLM vs. Human responses. The average score is computed assuming Level 1 to 5 correspond to scores from 1 to 5.}
\label{tab:human_agreement_comparison}
\end{table}

\section{Dataset Categories}
\label{sec:dataset_categories}

\begin{table}[H]
\centering
\small
\setlength{\tabcolsep}{3pt} 
\begin{tabular}{p{3.5cm}|c|c} 
\toprule
\textbf{Categories} & \textbf{MUN-vis} & \textbf{MUN-lang} \\
\midrule
\scriptsize Household Items and Furniture & 100 & 300 \\
\scriptsize Beverages & 82 & 22 \\
\scriptsize Fruits and Vegetables & 80 & 8 \\
\scriptsize Tools, Equipment & 57 & 143 \\ 
\scriptsize Dairy Products and Eggs & 54 & 0 \\
\scriptsize Health and Personal Care & 44 & 15 \\
\scriptsize Canned, Packaged, and Processed Goods & 36 & 5 \\
\scriptsize Meat and Seafood & 22 & 2 \\
\scriptsize Condiments and Sauces & 21 & 0 \\
\scriptsize Grains, Bread, and Baked Goods & 19 & 5 \\
\midrule
\textbf{Total} & \textbf{515} & \textbf{500} \\
\bottomrule
\end{tabular}

\vspace{-1mm}
\caption{Comparison of object category counts across textual description of visual context. Total counts for each dataset are provided in the last row.}
\label{tab:category_comparison}
\end{table}

\begin{figure*}[!h]
\footnotesize
    \centering
    \includegraphics[width=\linewidth]{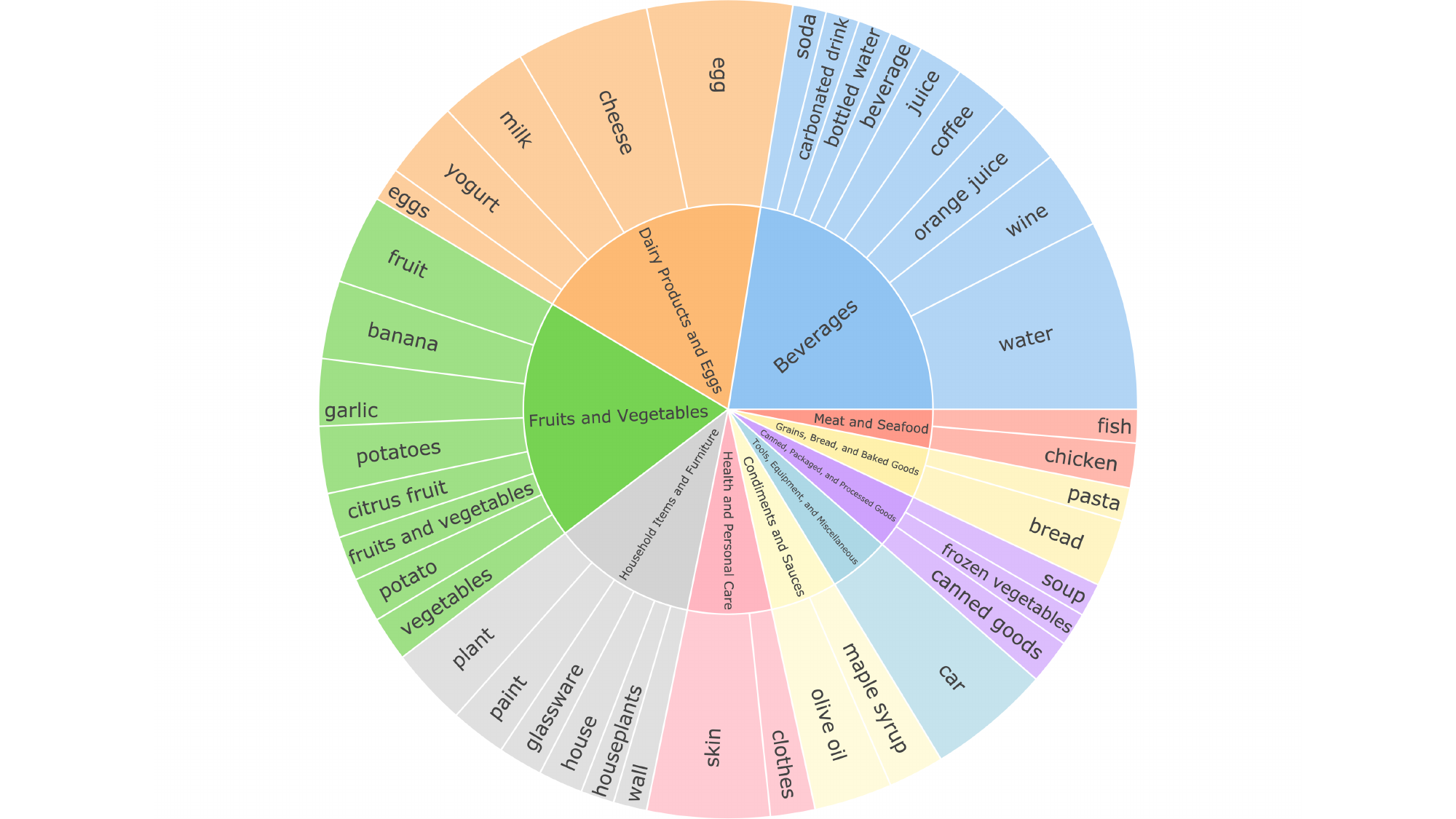}
    \vspace{1mm}
    
    \includegraphics[width=\linewidth]{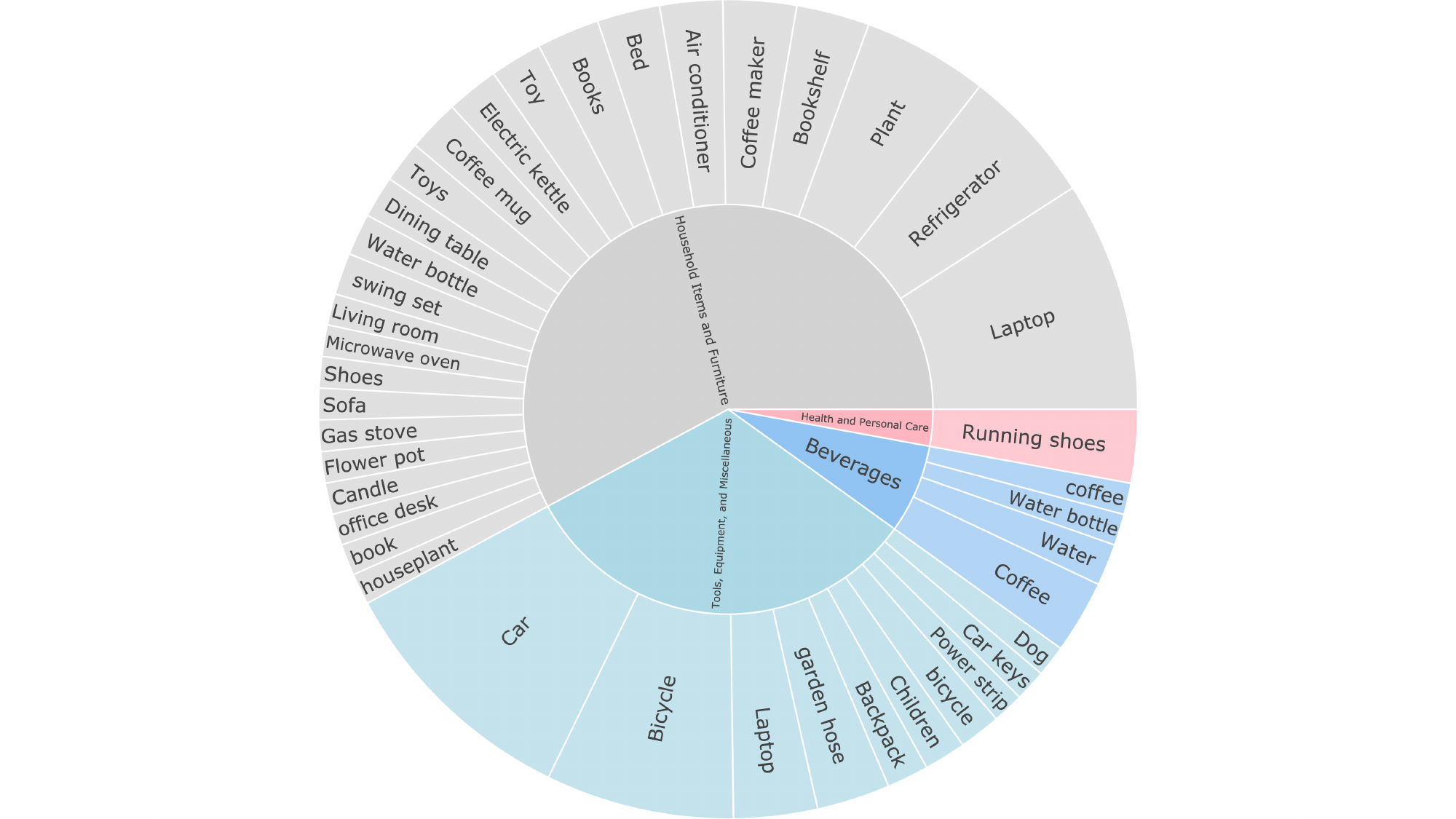}
    
    \caption{Textual context distribution in (top) MUN-vis and (bottom) MUN-lang.}
    \label{fig:sun_burst}
\end{figure*}

We selected the top 30 most frequent categories based on the textual context of MUN-vis and MUN-lang. As shown in Table \ref{tab:category_comparison}, MUN-vis focuses more on food-related elements, while MUN-lang emphasizes household and furniture items. However, the subsets still feature diverse subcategories and context-rich scenes at the example level, as illustrated in Figures \ref{fig:sun_burst}.

\clearpage
\section{Prompt used during Experiments}
\label{sec:prompts}
Figure \ref{tab:prompt_mun_vis} through \ref{tab:prompt_specificity} illustrate the various prompts used during dataset generation and evaluation.

\section{Human Annotation Details}

\subsection{Human Dataset Construction}

\begin{figure*}[!h]
\footnotesize
    \centering
    \includegraphics[width=\linewidth]{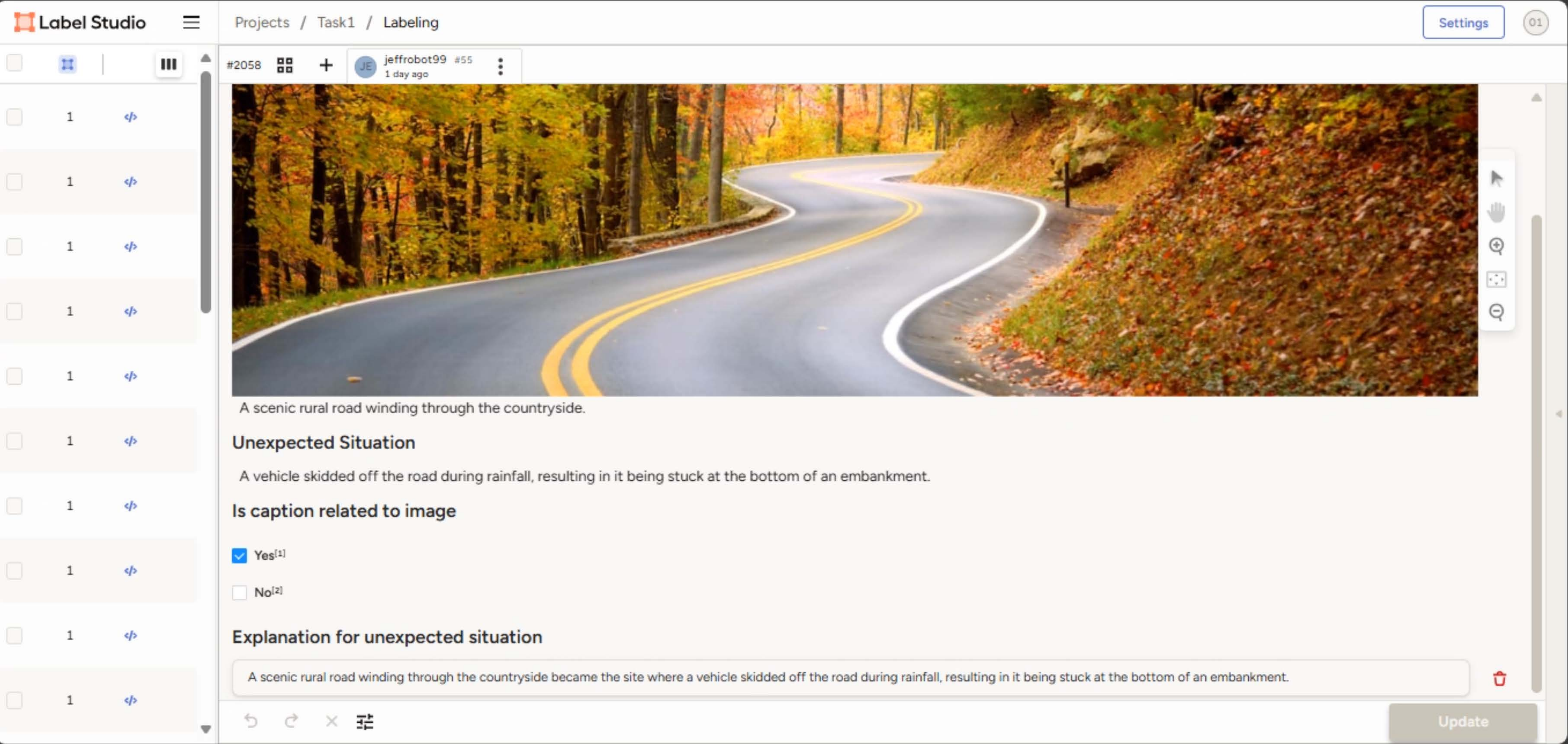}
    \caption{The user interface used for human annotation tasks, designed to facilitate the creation of detailed and contextually relevant explanations in MUN-vis and MUN-lang.}
    \label{fig:interface_human_explanation}

\end{figure*}

To construct the human-written dataset, we recruited 26 graduate students to generate contextualized explanations that logically bridge two provided segments. Annotators were instructed to follow a standardized interface that guided the construction of fluent and coherent connecting sentences. Each explanation was written with reference to the surrounding context to ensure narrative consistency. The interface used for collecting human explanations is illustrated in Figure~\ref{fig:interface_human_explanation}.

\subsection{Human Evaluation Protocol}
We conducted two human evaluation studies via the Prolific platform\footnote{\url{https://www.prolific.com}}, recruiting participants whose first language is English.

\begin{figure*}[!h]
\footnotesize
    \centering
    \includegraphics[width=0.7\linewidth]{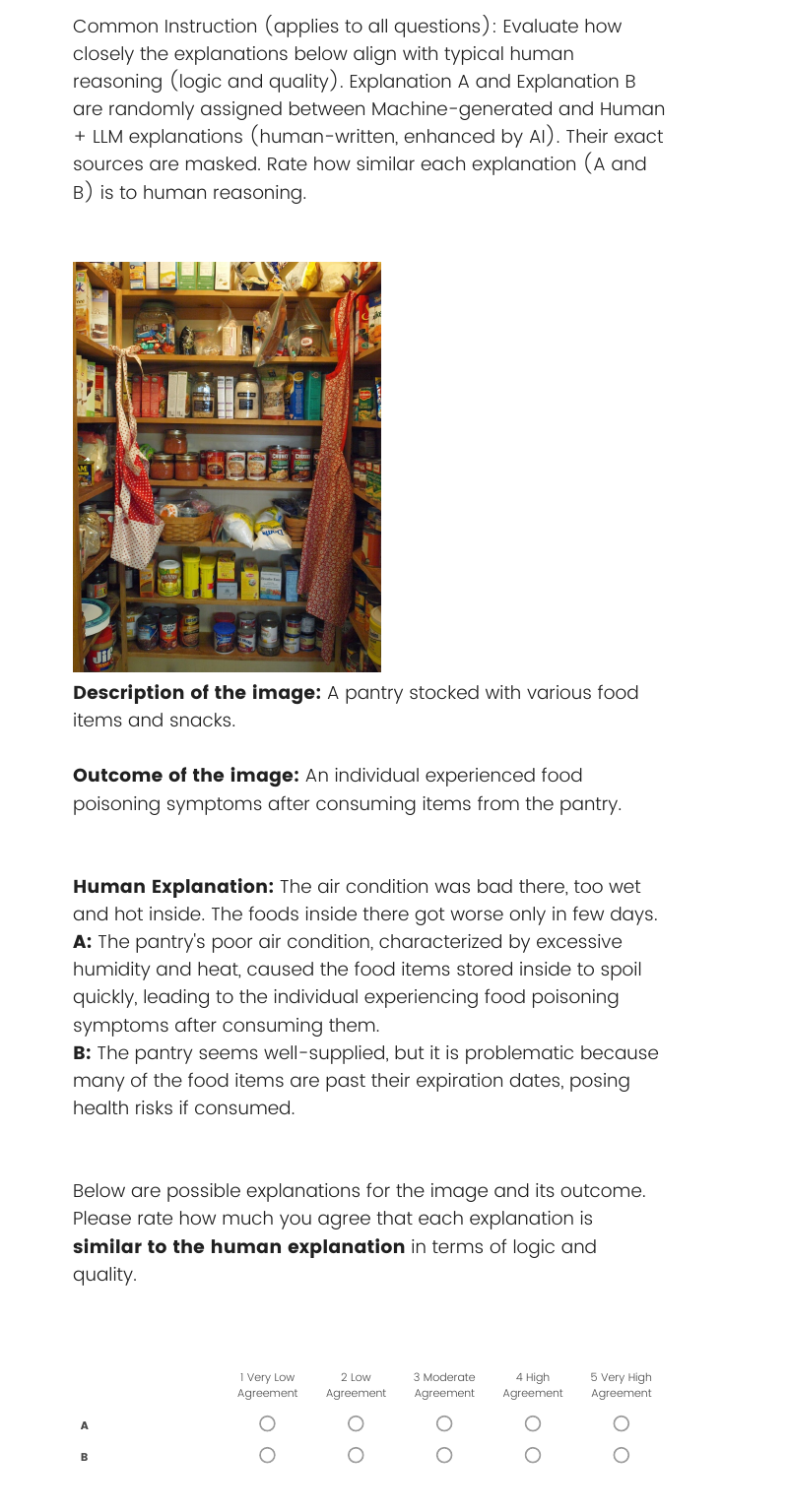}
    \caption{Interface used for human agreement evaluation. Annotators were presented with two anonymized responses, one written by a human, and the other either directly generated by an LLM or revised by an LLM based on the human version, and asked to select the more appropriate one.}

    \label{fig:human_comparison}
\end{figure*}

\paragraph{(1) Human Agreement Evaluation.}
To assess alignment between human and model-generated outputs, we asked annotators to compare two anonymized responses for each of 500 randomly selected samples, across two comparisons: (a) \textsc{LLM} vs. Human and (b) \textsc{HLLM} vs. Human. Each sample was evaluated by two independent annotators, resulting in a total of 2,000 judgments. A total of 141 unique participants were recruited for this task, and workers were compensated at a rate of €7.50 per hour.
The interface used for collecting human agreement on explanations is illustrated in Figure~\ref{fig:human_comparison}.

\begin{figure*}[!h]
\footnotesize
    \centering
    \includegraphics[width=0.7\linewidth]{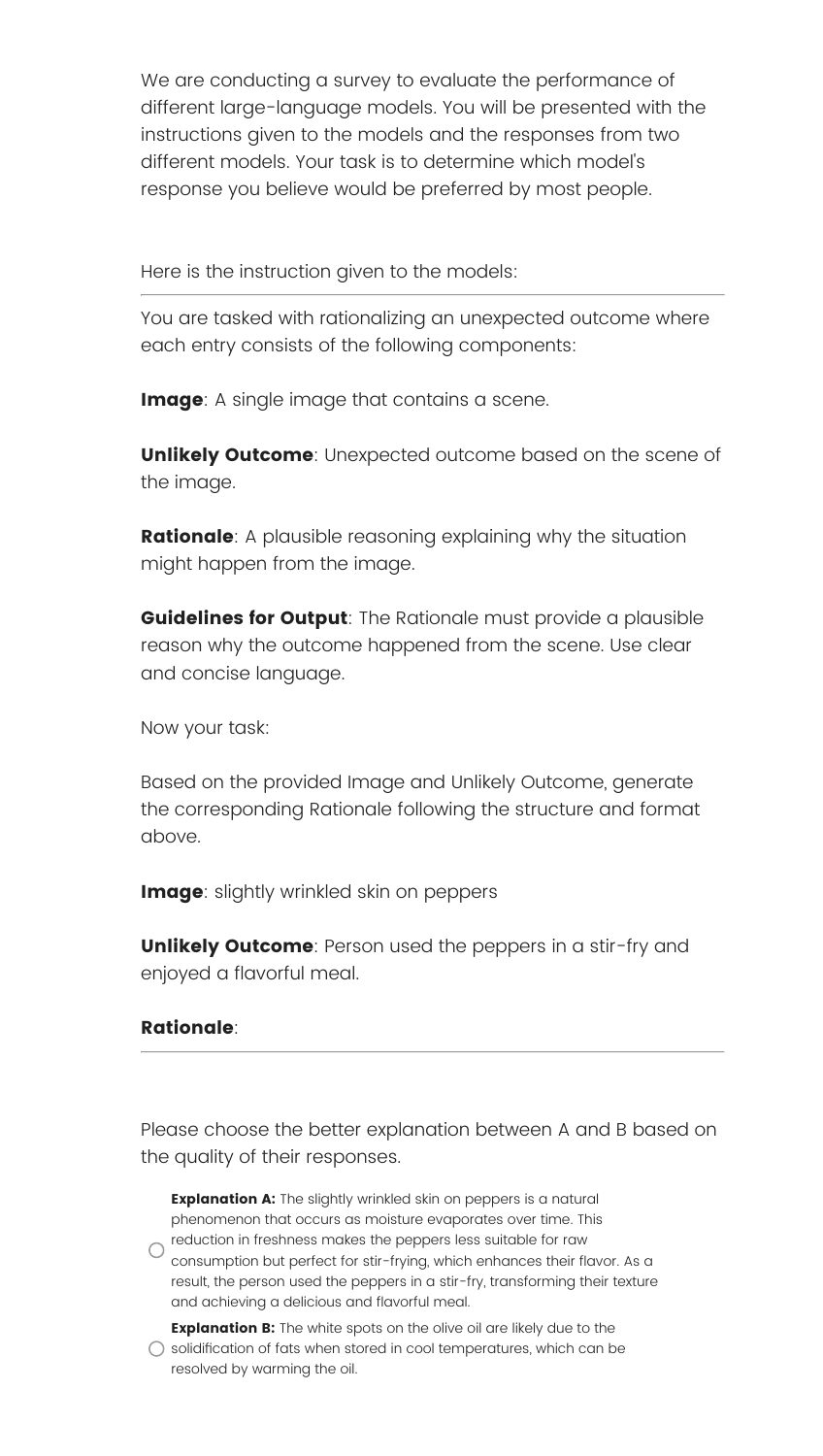}
    \caption{Interface presented to human annotators for evaluating pairwise win rates between model responses (e.g., zero-shot vs. retrieved 5-shot). Annotators were shown two anonymized outputs and asked to select the better one based on quality.}
    \label{fig:human_winrate}
\end{figure*}

\paragraph{(2) Win Rate Comparison.}
We further evaluated relative response quality across few-shot prompting variants (zero-shot, random 5-shot, retrieved 5-shot) using a win-rate setup. For each of 50 representative samples, we constructed 3 pairwise comparisons (e.g., retrieved vs. zero-shot), resulting in 150 comparisons per modality. This evaluation was conducted separately for \textsc{MUN-lang} and \textsc{MUN-vis}, yielding a total of 300 pairwise comparisons. Each comparison was rated by a single annotator. A total of 20 unique participants were recruited for this task, and they were compensated at a rate of €7.71 per hour.
The interface used for collecting win-rate judgments is shown in Figure~\ref{fig:human_winrate}.


\begingroup
\setlength{\textfloatsep}{8pt} 
\begin{figure*}[t]
\centering
\begin{minipage}{0.98\linewidth} 
\footnotesize 
\begin{tcolorbox}[
    colback=gray!5,
    colframe=gray!100,
    arc=2mm,           
    boxrule=1pt,       
    width=\linewidth,  
    left=1mm, right=1mm, top=1mm, bottom=1mm, 
    enhanced,
]
\begin{lstlisting}
You are tasked with generating a dataset where each entry consists of the following components:

Caption: A short description of an object or scene in an image.
Rationale: A plausible reasoning explaining why the object or scene might lead to an issue.
Situation: A potential outcome based on the caption and rationale, without explicitly mentioning the cause.

Guidelines for Output:
- The Situation must describe the outcome without directly linking it to the rationale.
- Use clear and concise language.
- Format the output for each entry as follows, enclosed in curly brackets {} to make it easy to parse:

{Caption: "<caption text>"} {Rationale: "<rationale text>"} {Situation: "<situation text>"}

Examples:

Example 1:
{Caption: "red liquid in steak packaging"} {Rationale: "The red liquid found in steak packaging is often mistaken for blood. It is actually a mixture of water and a protein called myoglobin that naturally occurs in muscle tissue. This liquid is perfectly normal and does not indicate that the meat is unsafe."} {Situation: "Person cooked and enjoyed the steak without health issues."}

Example 2:
{Caption: "settling of liquid in yogurt"} {Rationale: "When you open a container of yogurt, you might observe a layer of clear liquid on top, which some may believe signifies spoilage. This liquid is simply whey separating from the yogurt solids, a natural process that doesn't affect the yogurt's quality. Stirring the whey back into the yogurt will restore its creamy texture."} {Situation: "Person enjoyed the yogurt as part of their breakfast."}

Example 3:
{Caption: "green patina on copper cookware"} {Rationale: "Copper cookware may develop a greenish layer called patina. Some people mistake this for harmful corrosion, but patina is natural and can actually protect the copper from further oxidation. The cookware is still usable after proper cleaning."} {Situation: "Person used copper cookware to prepare a delicious meal."}

Example 4:
{Caption: "yellowing leaves on indoor plants"} {Rationale: "Indoor plant leaves may start to turn yellow as a natural part of their growth cycle or due to minor stress factors like overwatering. A few yellow leaves do not necessarily indicate that the plant is dying."} {Situation: "Person continued to care for the plant, and it grew healthy new leaves over time."}

Example 5:
{Caption: "skin peeling after a sunburn"} {Rationale: "After a sunburn, the skin may start to peel. This peeling is part of the natural healing process where the body sheds damaged skin cells. While it might look alarming, it is a normal response to skin damage from ultraviolet light exposure and not a cause for concern."} {Situation: "Person applied moisturizer and supported the skin's healing process comfortably."}

Now your task:
Based on the provided Caption and Rationale, generate the corresponding Situation following the structure and format above.

{Caption: "{INPUT CAPTION HERE}"}{Rationale: "{INPUT RATIONALE HERE}"}
\end{lstlisting}
\end{tcolorbox}
\vspace{-1.5mm} 
\caption{Prompt Template for Generating Scenarios for MUN-vis}
\label{tab:prompt_mun_vis}
\end{minipage}
\end{figure*}
\endgroup

\begin{figure*}[t]
\centering
\begin{tcolorbox}[
    colback=gray!5,
    colframe=gray!100,
    arc=3mm,
    boxrule=1.5pt,
    width=0.98\linewidth,
    left=2mm,
    right=2mm,
    top=1.5mm,
    bottom=1.5mm,
    enhanced,
]
\begin{lstlisting}
You are tasked with generating a dataset where each entry consists of the following components:

Caption: A short description of an object or scene in an image.
Rationale: A plausible reasoning explaining why the object or scene might lead to an issue.
Situation: A potential outcome based on the caption and rationale, without explicitly mentioning the cause.

Guidelines for Output:
- The Situation must describe the outcome without directly linking it to the rationale.
- Use clear and concise language.
- Format the output for each entry as follows, enclosed in curly brackets {} to make it easy to parse:

{Caption: "<caption text>"} {Rationale: "<rationale text>"} {Situation: "<situation text>"}

Examples:

Example 1:
{Caption: "A coffee maker ready to brew the perfect cup."} {Rationale: "While the coffee maker looks functional, its internals are corroded, leading to potential contamination of the brewed coffee."} {Situation: "A customer experienced stomach discomfort after drinking coffee brewed from the machine."}

Example 2:
{Caption: "A sleek sports car parked in the driveway."} {Rationale: "The sports car is problematic because it has an undiagnosed mechanical issue, making it dangerous to drive."} {Situation: "The driver encountered a sudden loss of control while driving, leading to a minor collision."}

Example 3:
{Caption: "A colorful toy ready for playtime."} {Rationale: "This is problematic because the toy is a recall item due to safety hazards that could pose a choking risk."} {Situation: "A child briefly choked while playing with the toy, requiring quick intervention."}

Example 4:
{Caption: "A desktop computer ready for work."} {Rationale: "The computer appears functional but is severely infected with malware that could compromise sensitive information."} {Situation: "The user faced unauthorized access to their private accounts after using the computer for online transactions."}

Now your task:
Based on the provided Caption and Rationale, generate the corresponding Situation following the structure and format above:

{Caption: "{INPUT CAPTION HERE}"}{Rationale: "{INPUT RATIONALE HERE}"}
\end{lstlisting}
\end{tcolorbox}
\caption{Prompt Template for Generating Scenarios for MUN-lang}
\label{tab:prompt_mun_lang}
\end{figure*}



\begin{figure*}[t]
\centering
\begin{tcolorbox}[
    colback=gray!5,
    colframe=gray!100,
    arc=3mm,
    boxrule=1.5pt,
    width=0.98\linewidth,
    left=2mm,
    right=2mm,
    top=1.5mm,
    bottom=1.5mm,
    enhanced,
]
\begin{lstlisting}
Can you improve this explanation so that it becomes more specific to the context and makes the outcome more likely to happen?

Context: {INPUT CONTEXT HERE}
Outcome: {INPUT OUTCOME HERE}
Explanation for the outcome: {INPUT EXPLANATION HERE}
\end{lstlisting}
\end{tcolorbox}
\caption{Prompt Template for improving the human explanation}
\label{tab:prompt_HLLM}
\end{figure*}

\begin{figure*}[t]
\centering
\begin{tcolorbox}[
    colback=gray!5,
    colframe=gray!100,
    arc=3mm,
    boxrule=1.5pt,
    width=0.98\linewidth,
    left=2mm,
    right=2mm,
    top=1.5mm,
    bottom=1.5mm,
    enhanced,
]

\textbf{System Prompt}

\begin{lstlisting}
You are a helpful assistant, that ranks models by the quality of their answers.

Prompt
I want you to create a leaderboard of different large-language models. To do so, I will give you the instructions (prompts) given to the models, and the responses of two models. Please rank the models based on which responses would be preferred by humans. All inputs and outputs should be Python dictionaries.

Here is the prompt:
{
    "instruction": """{instruction}"""
}

Here are the outputs of the models:
[
    {
        "model": "model_1",
        "answer": """{output_1}"""
    },
    {
        "model": "model_2",
        "answer": """{output_2}"""
    }
]

Now please rank the models by the quality of their answers, so that the model with rank 1 has the best output. Then return a list of the model names and ranks, i.e., produce the following output:
[
    {"model": "model_1", "rank": 1},
    {"model": "model_2", "rank": 2}
]

Your response must be a valid Python dictionary and should contain nothing else because we will directly execute it in Python. Please provide the ranking that the majority of humans would give.
\end{lstlisting}

\end{tcolorbox}
\caption{Prompt Template for Assessing Win Rate}
\label{tab:prompt_eval_winrate}
\end{figure*}
\begin{figure*}[t]
\centering
\begin{tcolorbox}[
    colback=gray!5,
    colframe=gray!100,
    arc=3mm,
    boxrule=1.5pt,
    width=0.98\linewidth,
    left=2mm,
    right=2mm,
    top=1.5mm,
    bottom=1.5mm,
    enhanced,
]
\begin{lstlisting}
You are tasked with evaluating the specificity of a given text on a scale of 1 to 5.
1 (Very Low Specificity): Extremely vague and general.
2 (Low Specificity): Limited details, mostly general.
3 (Moderate Specificity): Includes some details but still general in parts.
4 (High Specificity): Contains clear and detailed information.
5 (Very High Specificity): Extremely detailed and precise, leaving no room for ambiguity.

Only output the score as a single number.

Input Text:
[Insert the generated text here]

Output Format:
[Score (1-5)]
\end{lstlisting}
\end{tcolorbox}
\caption{Prompt Template for Assessing Specificity}
\label{tab:prompt_specificity}
\end{figure*}

\begin{figure*}[t]
\centering
\begin{tcolorbox}[
    colback=gray!5,
    colframe=gray!100,
    arc=3mm,
    boxrule=1.5pt,
    width=0.98\linewidth,
    left=2mm,
    right=2mm,
    top=1.5mm,
    bottom=1.5mm,
    enhanced,
]
\begin{lstlisting}
You are tasked with evaluating the specificity of a given text on a scale of 1 to 5.
1 (Very Low Specificity): Extremely vague and general.
2 (Low Specificity): Limited details, mostly general.
3 (Moderate Specificity): Includes some details but still general in parts.
4 (High Specificity): Contains clear and detailed information.
5 (Very High Specificity): Extremely detailed and precise, leaving no room for ambiguity.

Only output the score as a single number.

Input Text:
[Insert the generated text here]

Output Format:
[Score (1-5)]
\end{lstlisting}
\end{tcolorbox}
\caption{Prompt Template for Assessing Specificity}
\label{tab:prompt_specificity}
\end{figure*}

\end{document}